\documentclass[letterpaper,twocolumn,10pt]{article}
\PassOptionsToPackage{hyphens}{url}
\usepackage{usenix2019_v3}
\usepackage{amsmath,amssymb,amsfonts}
\usepackage{algorithmic}
\usepackage{graphicx}
\usepackage{textcomp}
\usepackage{tcolorbox}
\usepackage{booktabs}
\usepackage{multirow}
\usepackage{algorithm}
\newcommand{\challenge}[1]{\textbf{(#1)}}

\date{}
\title{\Large\bf AdaFloor: Adaptive Floor Planning and Rank-Time Reuse\\for Long-Tail MoE RL Rollouts}
\author{{\rm Jia Wei, Ziyi Qiu, Yi Xu, and Xuehai Qian}\\
{\normalsize\rm Department of Computer Science and Technology, Tsinghua University}\\
{\normalsize\rm Corresponding author: Xuehai Qian (\texttt{xuehaiq@tsinghua.edu.cn})}}
\hypersetup{pdftitle={AdaFloor: Adaptive Floor Planning and Rank-Time Reuse for Long-Tail MoE RL Rollouts},pdfauthor={Jia Wei, Ziyi Qiu, Yi Xu, Xuehai Qian},pdfsubject={Preprint}}
\newtcolorbox{insightbox}{
  colback=white,
  colframe=black,
  boxrule=0.8pt,
  arc=1.5mm,
  outer arc=1.5mm,
  boxsep=0pt,
  left=2mm,
  right=2mm,
  top=1.2mm,
  bottom=1.2mm,
  before skip=5pt,
  after skip=6pt,
  fontupper=\itshape
}

\newcommand{\insight}[2]{%
  \begin{insightbox}
  \textbf{$\bigstar$ Insight #1.} #2
  \end{insightbox}}

\newcommand{\ThroughputGain}{52.74}
\newcommand{\ResourceEfficiencyGain}{65.40}
\newcommand{\ThroughputGainLS}{8.57}
\newcommand{\EfficiencyGainLS}{17.56}

\newcommand{\ResourceCostReduction}{39.5}
\newcommand{\TGTPSGain}{3.65}

\newcommand{\TGResourceGain}{1.68}

\newcommand{\QuotaQGain}{1.71}
\newcommand{\QuotaEGain}{7.47}
\newcommand{\QuotaWaitReduction}{61.26}
\newcommand{\GraphBaselineGain}{90.76}
\newcommand{\GraphPfourGain}{124.88}

\newcommand{\CodeQGain}{22.02}

\newcommand{\CodeEGain}{30.35}

\begin{document}
\setlength{\textfloatsep}{10pt plus 2pt minus 2pt}
\setlength{\dbltextfloatsep}{10pt plus 2pt minus 2pt}
\setlength{\floatsep}{8pt plus 2pt minus 2pt}
\setlength{\dblfloatsep}{8pt plus 2pt minus 2pt}
\setlength{\intextsep}{10pt plus 2pt minus 2pt}
\setlength{\abovecaptionskip}{5pt}

\newlength{\savedFirstPageFloatSep}
\setlength{\savedFirstPageFloatSep}{\textfloatsep}
\setlength{\textfloatsep}{12pt}
\AddToHook{build/column/before}[first-page-float-gap]{%
  \ifnum\value{page}=1\relax\raggedbottom\fi}
\AddToHookNext{shipout/after}{%
  \global\textfloatsep=\savedFirstPageFloatSep\relax
  \RemoveFromHook{build/column/before}[first-page-float-gap]}
\maketitle

\begin{abstract}
Reinforcement learning post-training improves the capabilities of large
language models, but generating training responses (rollouts) is a major computational
bottleneck. In synchronous rollouts for mixture-of-experts models, a few long
responses can keep an entire expert-parallel group occupied. Because experts
are distributed across accelerators, devices that finish their local rollouts
must continue serving expert requests from devices that are still generating.
In our measurements, this interval accounts for approximately one-third of
total allocated accelerator time during rollouts.
We present AdaFloor, a system that shortens these intervals and releases
accelerators early during synchronous rollouts. AdaFloor combines an offline
planner with a graph-enabled runtime executor. The planner uses historical
rollout lengths to form batches and set output-length limits. It coordinates
prompt placement so enough accelerators finish early to leave the group
together, while selecting group sizes that fit device memory. The executor
monitors completion, releases finished accelerators for other tasks while
preserving expert availability and communication correctness, and returns them
to the group before policy updates. Compared with the verl baseline, AdaFloor
improves rank-time resource efficiency by approximately 65.40\% and rollout
throughput by 52.74\%, with little observed variation in rollout reward.
By making reinforcement learning
post-training more efficient, AdaFloor supports faster development and
improvement of large language models and artificial intelligence agents.
\end{abstract}

\section{Introduction}
\label{sec:intro}

Reinforcement learning (RL) post-training improves reasoning and instruction
following in large language models~\cite{ziegler2019fine, stiennon2020learning,ouyang2022training,shao2024deepseekmath,guo2025deepseekr1}.
Each RL post-training iteration generates responses with the current policy, computes
rewards for those responses, and updates the policy. The response-generation
phase, or \emph{rollout}, is a major performance bottleneck in RL
post-training~\cite{gao2026rollpacker}.
For example, prior measurements of reasoning RL report that rollout accounts
for approximately 85\% of total RL post-training step time~\cite{hu2026tlt}.

We focus on synchronous RL post-training, which maintains on-policy sampling
without introducing policy staleness or changing the policy-update
protocol~\cite{hu2026tlt}. However, each policy update must wait for the entire
rollout batch, so a few long responses delay progress even after most
accelerators finish their local rollouts.

\begin{figure}[t]
 \centering
 \setlength{\abovecaptionskip}{1pt}
 \includegraphics[width=0.92\columnwidth]{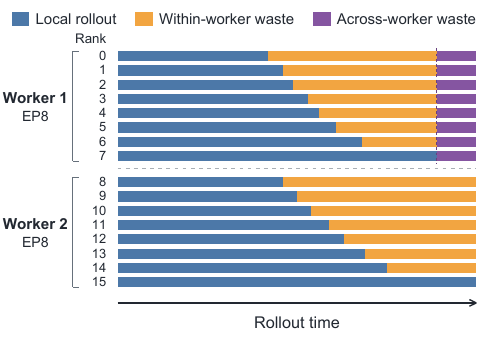}
 \caption{Within-worker and across-worker waste (schematic). Yellow marks
 locally finished ranks in active workers. Purple marks a finished worker
 waiting for the other.}
 \label{fig:dummy_timeline}
\end{figure}

This wait has an additional cost for sparse mixture-of-experts (MoE) models,
which increase capacity by activating only a subset of experts per token~\cite{shazeer2017sparsely,lepikhin2021gshard,fedus2022switch, du2022glam,jiang2024mixtral,dai2024deepseekmoe}.
Expert parallelism (EP) distributes weights to reduce per-device memory use
and exploit aggregate memory bandwidth~\cite{rajbhandari2022deepspeedmoe,hwang2023tutel,deepseekai2025deepseekv3}.
An EP group forms a \emph{rollout worker} serving one model replica, with one
process, or \emph{rank}, per accelerator. A locally finished rank still serves
expert requests and joins collectives until its worker finishes~\cite{lepikhin2021gshard,rajbhandari2022deepspeedmoe,hwang2023tutel,gale2023megablocks}.
During this \emph{dummy run}, attention is skipped or run on dummy tokens to
match the graph's input shape. The accelerator stays occupied without advancing
local rollouts. Dummy run accounts for 32.41\% to 33.39\% of total allocated
accelerator time in our verl baseline~\cite{sheng2025hybridflow}.
Figure~\ref{fig:dummy_timeline} shows dummy runs in yellow.

{\clubpenalty=10000
Prior work speeds up rollouts or reduces and reuses cross-worker idle time (purple). Seer combines fine-grained
scheduling with speculative decoding~\cite{qin2026seer}, while
Distribution-Aware Speculative Decoding (DAS) gives longer responses larger
draft budgets~\cite{shao2026das}. RollPacker reduces waiting across workers by
grouping long responses into dedicated rounds~\cite{gao2026rollpacker}.
\mbox{\emph{Taming the Long-Tail}} (TLT) uses idle workers to train a
drafter~\cite{hu2026tlt}. None releases ranks
from an active EP group to reclaim the yellow intervals, which occur even
when a single EP group spans all accelerators. We propose
AdaFloor to shorten dummy runs in these intervals and release locally finished
accelerators early for other tasks. Its batching and response limits also shorten
the purple interval when workers finish closer together. Safe release poses
three challenges. \challenge{1} Remaining ranks need memory for expert weights,
key-value (KV) caches, and workspace. \challenge{2} Enough ranks must finish
early to form a smaller EP group. \challenge{3} Expert availability and
collective ordering must hold as ranks leave and rejoin.\par}

\begin{figure*}[t]
  \centering
  \includegraphics[width=0.96\textwidth]{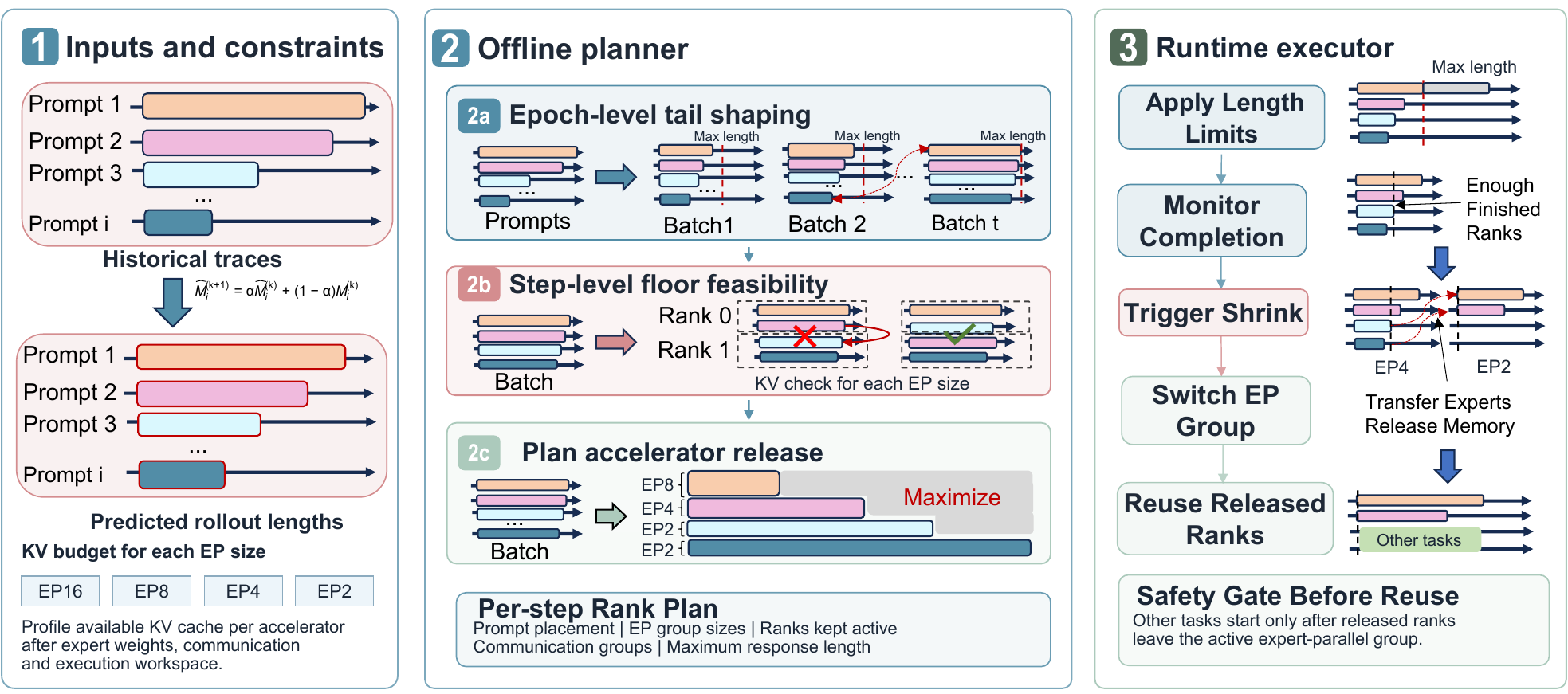}
  \caption{AdaFloor combines an offline planner with a runtime executor.
  The planner forms batches, sets output-length limits, and jointly selects
  prompt placement and EP group sizes under memory constraints. The executor
  applies the plan, releases finished ranks for other tasks, and restores
  the full group before policy updates.}
  \label{fig:adafloor_overview}
\end{figure*}

AdaFloor addresses these challenges with a co-designed \emph{offline planner}
and \emph{runtime executor}, shown in Figure~\ref{fig:adafloor_overview}.
The offline planner uses historical response lengths to form batches and set
output-length limits. It coordinates prompt placement so that enough ranks
finish early to leave the group together, while selecting EP group sizes
that fit the remaining accelerators' memory. These decisions address
challenges \challenge{1} and \challenge{2}. The runtime executor monitors
actual completion, releases finished ranks for other tasks while preserving
expert availability and collective ordering, and restores the full group
before policy updates, addressing challenge \challenge{3}.

On the DeepScaleR dataset~\cite{tan2025deepscaler}, AdaFloor improves rollout
throughput by 52.74\% and rank-time resource efficiency by approximately
65.40\% over the verl baseline.
Rollout reward shows little observed variation. On the Eurus-2-RL dataset~\cite{cui2025prime}, a matched
frozen-policy comparison improves throughput by 22.02\% and resource efficiency
by approximately 30.35\%. During post-training, released accelerators also run
inference with \mbox{Qwen2.5-1.5B}, \mbox{Qwen3-8B, and PanGu-MoE.}
Using RollPacker in the first epoch and AdaFloor thereafter yields higher
cold-start rollout throughput than either method alone.

\noindent\textbf{This paper makes the following contributions.}
\begin{itemize}
  \item We identify and quantify dummy-run waste caused by expert dependencies
  within active EP groups, showing why local rollout completion does not
  release accelerators for other tasks.
  \item We design an \emph{offline planner} that uses rollout-length history
  to coordinate batching, output-length limits, prompt placement, and EP group
  sizes under device memory constraints, enabling earlier release
  and shorter rollouts.
  \item We implement a graph-enabled \emph{runtime executor} that preserves
  expert availability and collective ordering as ranks leave and rejoin.
  Experimental results show throughput and efficiency gains and confirm reuse for inference.
\end{itemize}

\section{Background}
\label{sec:background}

\subsection{RL Post-Training Workflow}
\label{sec:bg:rollout}

RL post-training optimizes a language model using rewards from its generated
responses. Common algorithms include proximal policy optimization (PPO),
group relative policy optimization (GRPO), and decoupled clip and dynamic
sampling policy optimization (DAPO)~\cite{schulman2017ppo,shao2024deepseekmath,yu2025dapo}.
They differ in their update rules but all generate responses autoregressively
during rollout, compute rewards, and update the policy.
Rollout often dominates step time~\cite{gao2026rollpacker,hu2026tlt}.

Asynchronous execution overlaps rollouts with policy updates to keep devices
busy. If updates outpace generation, the learner must account for samples from
older policies~\cite{fu2025areal}. Synchronous execution generates a complete
batch with the current policy before updating it, preserving on-policy sampling
without additional policy lag~\cite{hu2026tlt}. We target accelerator waste
within this synchronous workflow.

\subsection{Data and Expert Parallelism in MoE}
\label{sec:bg:ep}

Combining data parallelism (DP) for attention with EP for experts, a
configuration we call DPEP, is a widely adopted deployment pattern for
large-scale MoE inference~\cite{deepseekai2025deepseekv3,zhu2025megascaleinfer,wang2026ainferpd}.
In DPEP rollouts, ranks process different requests using replicated nonexpert
parameters, while expert weights are distributed across the same
ranks~\cite{rajbhandari2022deepspeedmoe,hwang2023tutel}. At each MoE layer, the
router selects experts, token dispatch sends tokens to the corresponding ranks
for expert computation, and combine returns weighted results to the request's
owning rank. Dispatch and combine use all-to-all communication across the EP
group~\cite{lepikhin2021gshard,rajbhandari2022deepspeedmoe}.

\begin{figure}[t]
  \centering
  \includegraphics[width=0.92\columnwidth]{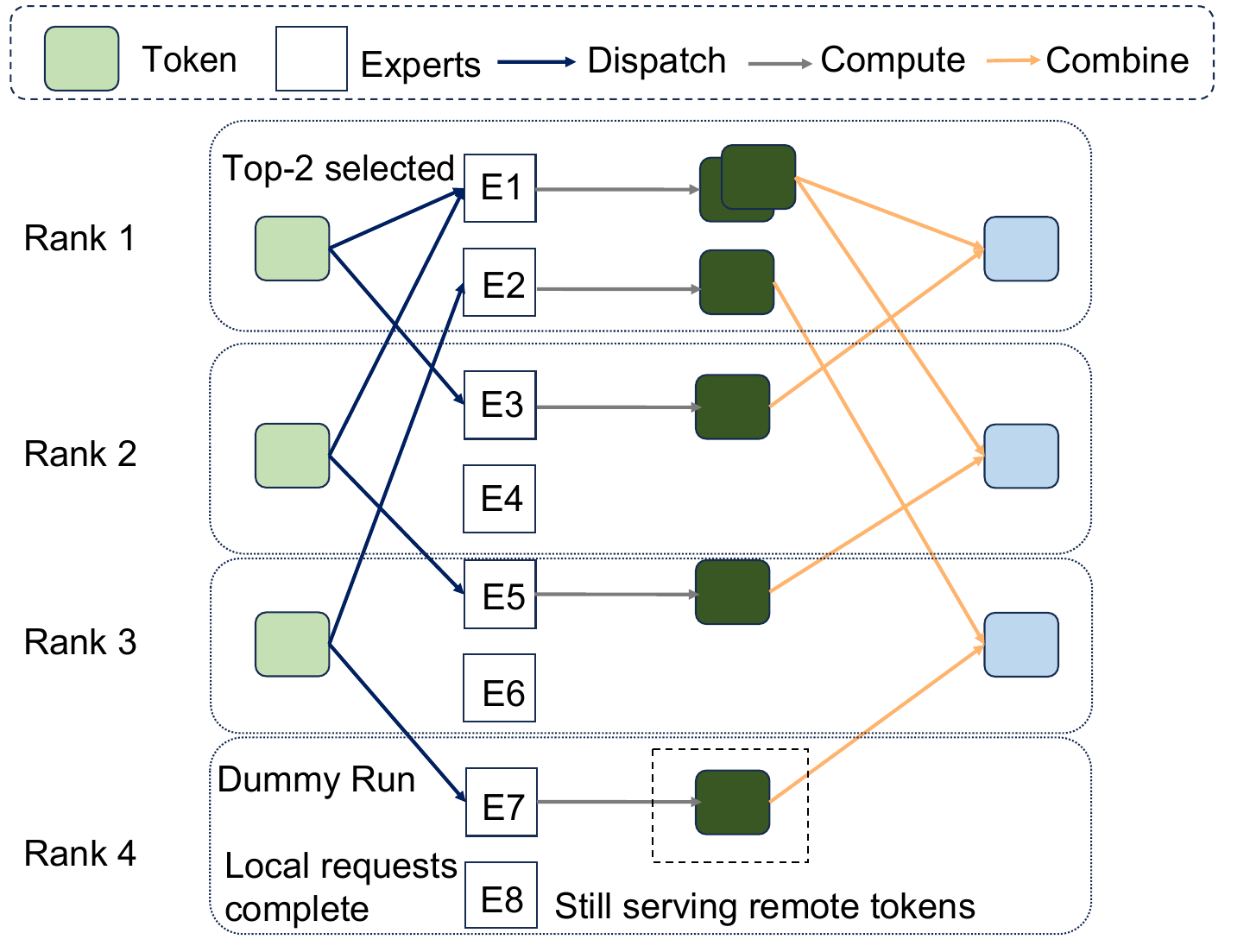}
  \caption{Token routing in a DPEP group. The dashed box marks expert
  computation during Rank 4's dummy run.}
  \label{fig:ep_coupling}
\end{figure}

Figure~\ref{fig:ep_coupling} shows the path of a token owned by Rank 3. Rank 3
dispatches the token to expert E2 on Rank 1 and expert E7 on Rank 4, then
combines their outputs. Rank 4 computes its output even though its own requests
have finished.

\section{Characterizing Long-Tail Waste}
\label{sec:characterization}

\subsection{Methodology}
\label{sec:char:methodology}

We analyze 30 training steps of the verl baseline using Qwen3-30B-A3B on
DeepScaleR. The runs use one EP group spanning 16 Ascend devices and cover
two epochs for each of three seeds. Each rollout step generates 16 responses
per prompt for 32 prompts, with 32 responses assigned to each rank.

\subsection{\textbf{Observation 1. Ranks Stay Allocated}}
\label{sec:char:tail}

In the verl baseline, 32.41\% to 33.39\% of the total accelerator time
allocated to rollout occurs after ranks finish their own requests but before
the group finishes. Figure~\ref{fig:dummy_timeline} illustrates this interval
in yellow.

\insight{1}{Fixed-group DPEP rollouts can tie up substantial accelerator time
after ranks finish their own requests, leaving those devices unavailable for
other tasks.}

\subsection{\textbf{Observation 2. History Predicts Length}}
\label{sec:char:prediction}

\begin{figure}[t]
 \centering
 \setlength{\abovecaptionskip}{1pt}
 \includegraphics[width=0.96\columnwidth,trim=0 2bp 0 0,clip]{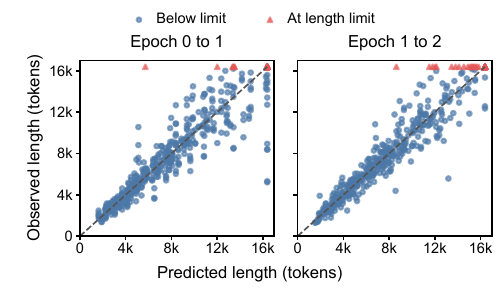}\par
 \caption{Predicted versus observed lengths. The diagonal marks
 exact agreement and triangles mark length-limit hits.}
 \label{fig:prompt_max_adjacent_epochs}
\end{figure}

We predict each prompt's longest response from its historical lengths.
Figure~\ref{fig:prompt_max_adjacent_epochs} shows 960 comparisons across 160
prompts, three seeds, and two epochs. Of the 846 comparisons in which neither
length hit the limit, 89.1\% of predictions fall within 25\% of the observed
length. In 7 cases (0.8\%), the observed length exceeds the prediction by
more than 50\%. Although rare, such underestimates can prolong synchronous
rollout because the batch must wait for every response to finish.

\insight{2}{A prompt's past response lengths help predict its future lengths,
but the planner must limit the impact of unusually long responses.}

\subsection{\textbf{Observation 3. Pairing, KV, and Release}}
\label{sec:char:kv}

\begin{figure}[t]
  \centering
  \includegraphics[width=\columnwidth]{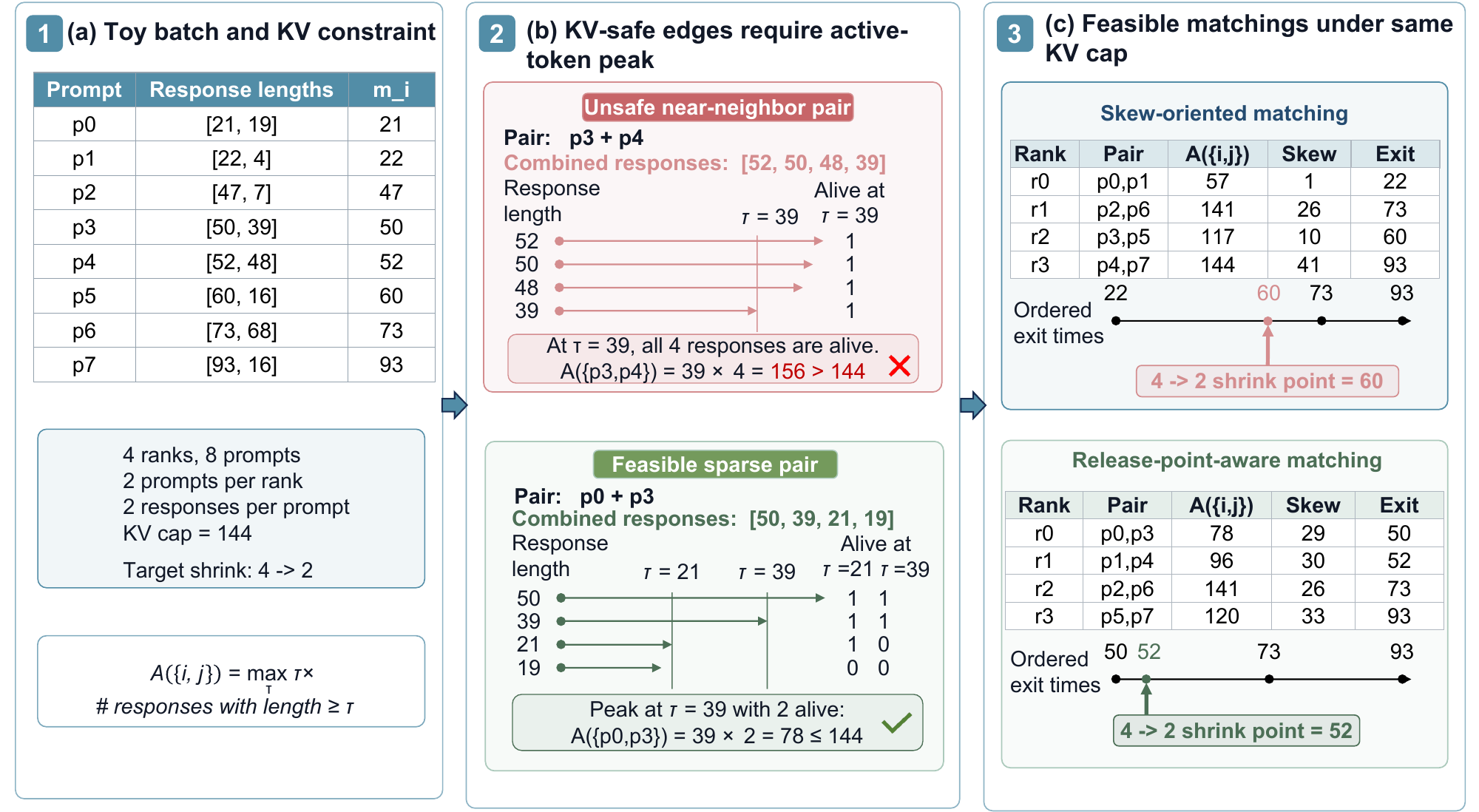}
  \caption{Prompt pairing on four ranks, each with capacity for 144 cached
  tokens. For prompt $p_i$, $m_i$ is its longest response length. The figure
  uses $\tau$ for the decoding step and $A(\{i,j\})$ for the peak number of
  cached tokens for prompts $p_i$ and $p_j$. Skew is the difference between
  their longest response lengths. Exit marks local completion.}
  \label{fig:toy_kv_matching}
\end{figure}

When two prompts share a rank, their responses compete for KV cache space.
Even prompts with similar maximum response lengths can exceed the cache
capacity when paired.
In Figure~\ref{fig:toy_kv_matching}, all four responses from $p_3$ and $p_4$
are still active at decoding step 39. They require $4\times39=156$ cached
tokens, more than the rank's capacity of 144. Pairing $p_3$ with $p_0$
instead leaves only two responses active at that step and peaks at 78 tokens.

Fitting in memory is not enough for early release. To shrink the group from
four ranks to two, at least two ranks must finish their local responses. In
the figure, pairing prompts to reduce the difference between their longest
response lengths lets the first rank finish at step 22, but the second does
not finish until step 60. An alternative pairing has two ranks finish by
step 52 without delaying the end of the batch. Both assignments fit in
memory, but the alternative releases two ranks eight decoding steps earlier.

\insight{3}{Prompt pairing must account for overlapping responses to fit in
memory and help enough ranks finish early to release them together.}

\subsection{\textbf{Observation 4. EP Size Varies by Step}}
\label{sec:char:floor}

In one measured epoch, the EP group shrank to two ranks in the first step
but stayed above eight in the last, when at least nine ranks remained busy.
\emph{Released rank-time} is the sum of the intervals from each rank's
release until the primary job reclaims it.

The smallest feasible EP group does not always maximize released rank-time.
In an offline replay of a recorded step, both EP2 and EP4 fit in memory.
The tighter KV budget of EP2 delayed the first eight releases from decode
step 3,406 to 4,426. Two more ranks left at step 5,488, too late to make up
for that delay. The resulting release was 25,718 rank-token units for EP2
versus 32,364 for EP4.

\insight{4}{EP group size is a step decision. AdaFloor compares the best memory-safe schedule for each supported size and selects the one with the most predicted released rank-time.}

\subsection{\textbf{Observation 5. Reusing Released Ranks}}
\label{sec:char:reuse}

In our experiments, released accelerators served inference requests for
three models without adding devices or substantially slowing the primary
rollout. They then rejoined the EP group before policy updates
(Section~\ref{sec:eval:reuse}).

\insight{5}{Ranks released from an active EP group can run other tasks
until the primary job needs them back.}

\section{AdaFloor Overview}
\label{sec:overview}

AdaFloor combines an offline planner with a runtime executor, as shown in
Figure~\ref{fig:adafloor_overview}. The planner uses historical rollout lengths
and memory profiles to form batches, set output-length limits, and plan prompt
placement and EP group sizes. The executor monitors actual completion and
safely releases accelerators for other tasks before returning them to the group.
This section introduces the inputs and constraints
(Section~\ref{sec:overview:inputs}), EP group sizes and completion quorums
(Section~\ref{sec:overview:groups}), and the planning objective and component
flow (Section~\ref{sec:overview:objective}). Sections~\ref{sec:epoch}
and~\ref{sec:floorplanning} detail epoch-level batching and per-step release
planning. Section~\ref{sec:runtime} describes the executor, and
Section~\ref{sec:evaluation} reports the evaluation.

\subsection{Inputs and Constraints}
\label{sec:overview:inputs}

The first input is the prompt set and its historical rollout lengths. For each
prompt, AdaFloor retains both its maximum response length and the full multiset
of sampled lengths. The maximum predicts when its assigned rank will finish.
The multiset estimates how many responses remain active together and therefore
how much KV cache they require. AdaFloor records rollout lengths in an initial
epoch without shrinking and refreshes the records after each subsequent epoch.

The second input is the set of EP group sizes supported by the runtime and a
memory profile for each size and execution configuration. We denote the usable KV
token capacity per accelerator by $C_{f,m}^{\mathrm{eff}}$, where $f$ is the
EP group size and $m$ identifies the runtime profile, including its residency
policy. The budget counts the tokens in whole KV cache blocks that fit in
the high-bandwidth memory (HBM) left after reserving space for weights,
communication buffers, workspace, and prompt KV.
Section~\ref{sec:eval:setup} gives the evaluated capacities.

The planner also takes the number of prompts per step $B$, the number of
responses per prompt $n$, and the global output-length limit as inputs.
It assigns each prompt to exactly one rank. At each transition, the
remaining ranks must have enough KV cache for their active responses.
If no smaller group fits, execution keeps the original group.

\subsection{EP Group Sizes and Completion Quorums}
\label{sec:overview:groups}
\label{sec:bg:floors}

We call the number of ranks retained after a transition a \emph{shrink floor},
or simply a floor. A transition from $N$ ranks to $f$ ranks requires at least
$N-f$ ranks to finish their local responses. This completion threshold is a
\emph{quorum}. A single early completion is not enough when the next
supported group size requires several ranks to leave together.

The evaluated runtime uses progressive halving. For example, the sequence
$16{\rightarrow}8{\rightarrow}4{\rightarrow}2$ becomes possible after 8,
12, and 14 ranks complete. Selecting a final group of two ranks includes the
earlier transitions through eight and four ranks. It does not postpone all
release until the last transition. Conversely, a smaller final group is not
always preferable if its tighter memory budget delays earlier exits.

\subsection{Planning Objective and Component Flow}
\label{sec:overview:objective}

AdaFloor first minimizes predicted epoch rollout time and then maximizes
predicted released rank-time within the resulting safe layouts. Let $\pi$
denote an epoch plan produced by sorted batching and bounded repair. Let $\Pi_{\mathrm{safe}}$
contain the candidates satisfying runtime legality, KV capacity, remaining-rank
memory, and output-length constraints. AdaFloor selects
\begin{equation}
\pi^\star=\operatorname*{lexmin}_{\pi\in\Pi_{\mathrm{safe}}}
\left(\widehat T_{\mathrm{epoch}}(\pi),-\widehat R(\pi)\right).
\label{eq:lexicographic_objective}
\end{equation}
Here $\widehat T_{\mathrm{epoch}}$ is the predicted epoch rollout time and
$\widehat R$ is the predicted released rank-time summed across steps.
For a fixed batch and output-length limit, prompt placement determines the
predicted completion times of individual ranks without changing the batch's
predicted completion time.

Figure~\ref{fig:adafloor_overview} shows how the planner implements this
objective. The epoch planner proposes batches and output-length limits
(panel 2a). For each candidate batch, the per-step planner checks memory
feasibility and decides when and how many ranks to release (panels 2b and 2c).

\section{Offline Planner for Epoch Batches}
\label{sec:epoch}

The epoch planner groups prompts into rollout batches and sets an output-length
limit for each batch (Figure~\ref{fig:adafloor_overview}, panel 2a). It uses the
per-step planner (Section~\ref{sec:floorplanning}) to validate candidate batches
and plan prompt placement and rank release.

\subsection{Predict Lengths and Form Batches}
\label{sec:epoch:prediction}

For prompt $i$ in epoch $k$, let $L_{i,j}^{(k)}$ be the generated length of its
$j$-th sampled response. AdaFloor defines the realized prompt tail as
\begin{equation}
M_i^{(k)}=\max_{j\in[1,n]}L_{i,j}^{(k)}.
\label{eq:prompt_tail}
\end{equation}
Using the maximum matches the synchronized objective. One long sample is enough
to delay the prompt's owning rank. AdaFloor updates each prompt's prediction
using an exponential moving average~\cite{roberts1959geometric}
\begin{equation}
\hat M_i^{(k+1)}=
\alpha\hat M_i^{(k)}+(1-\alpha)M_i^{(k)}.
\label{eq:ema}
\end{equation}
The smoothing factor $\alpha$ balances the historical estimate and the most
recent observation. Section~\ref{sec:eval:setup} gives the evaluated setting.
The complete response-length multiset is retained separately for KV estimation.
The predictor is maintained at dataset-item granularity because prompts sharing
a template can still have different reasoning depth. The initial no-shrink epoch
provides the first observation, and each completed epoch updates the estimate
used to plan the next one.

AdaFloor sorts the epoch prompts by $\hat M_i$ and partitions the sorted order
into contiguous $B$-prompt steps. Let $P_{\mathrm{sorted}}$ be this order. Step
$t$ receives
\begin{equation}
B_t=\{P_{\mathrm{sorted}}[r]\mid (t-1)B\le r<tB\}.
\end{equation}
The longest predicted response sets each step's estimated duration. Sorting
groups prompts with long predicted responses into fewer batches, minimizing
the sum of estimated step durations for a fixed batch size.

\subsection{Set Per-Batch Output-Length Limits}
\label{sec:epoch:tailguard}

One unexpectedly long response can keep a short batch running up to the
global output-length limit. AdaFloor sets a limit for each batch to curb
such delays while leaving room for prediction error. For each prompt $i$
observed in epoch $k$, AdaFloor compares its predicted maximum response
length $\hat M_i^{(k)}$ with the observed maximum $M_i^{(k)}$ across that
prompt's responses. This gives one ratio per prompt per epoch
\begin{equation}
r_{i,k}^+=\max\left(1,\frac{M_i^{(k)}}{\max(1,\hat M_i^{(k)})}\right).
\label{eq:under_ratio}
\end{equation}
Overestimates give a ratio of one because they need no extra tokens.
Let $\mathcal W$ contain the most recent
$K$ epochs with comparisons and $\mathcal I_k$ be the prompts with both a
prediction and an observation in epoch $k$. AdaFloor pools these ratios
and takes their $q$-quantile as the multiplier
\[
\rho_q=\operatorname{Quantile}_q\!\left(
  \bigl(r_{i,k}^+\bigr)_{k\in\mathcal W,\,i\in\mathcal I_k}
\right).
\]
This is the cutoff for the lowest fraction $q$ of the pooled ratios. With
no historical comparisons, AdaFloor uses a configured default multiplier.

For a new batch $B_t$, AdaFloor starts with the longest predicted response
length $S_t=\max_{i\in B_t}\hat M_i$. Multiplying $S_t$ by $\rho_q$ adds
room for underestimation seen in prior epochs. AdaFloor bounds the scaled
estimate by the global output-length limit $L_{\max}$ and a configured
minimum $C_{\min}$ that prevents overly tight limits for short predictions.
The batch limit is
\begin{equation}
C_t=\min\left(L_{\max},\max\left(C_{\min},\rho_q S_t\right)\right).
\label{eq:tailguard}
\end{equation}
The executor truncates any unfinished response at $C_t$ tokens.
Figure~\ref{fig:adafloor_overview} (panel 2a) shows these per-batch limits.
Section~\ref{sec:eval:setup} lists the evaluated parameters.

\subsection{Batch Validation and Bounded Repair}
\label{sec:epoch:validation}

Let $\mathcal{P}$ denote the per-step planner described in
Section~\ref{sec:floorplanning}. The epoch planner passes it a candidate
batch $B_t$ and its output-length limit $C_t$
\begin{equation}
\mathcal{P}(B_t,C_t)\rightarrow(\mathrm{success},\phi_t),
\label{eq:planner_interface}
\end{equation}
On success, $\phi_t$ specifies prompt placement and the planned changes to
the EP group. When $\mathcal{P}$ finds a feasible plan, the epoch planner
keeps the batch and saves its output-length limit and execution plan for
direct use by the runtime executor.

If $\mathcal{P}$ fails under every allowed floor, AdaFloor performs bounded
repair with the preceding step and reruns $\mathcal{P}$ on the affected steps.
Starting from the last rejected step, it exchanges that step's shortest prompt
with the preceding step's shortest prompt. If validation still fails, it
re-sorts the preceding step and advances to the next-shortest prompt in the
rejected step. Each swap must lower the rejected step's KV demand. This bounded
search keeps the perturbation local and prioritizes swaps with the smallest
change to adjacent step maxima.

Algorithm~\ref{alg:epoch_planner} summarizes epoch construction and per-step
plan validation.

\begin{algorithm}[t]
  \caption{Epoch construction from historical response traces. Every retained
  step has a validated floor, quorum schedule, and prompt-to-rank assignment.}
  \label{alg:epoch_planner}
  \begin{algorithmic}[1]
    \REQUIRE Epoch prompts and historical response traces
    \ENSURE Planner validated epoch plan $\Pi$

    \STATE Update tail predictions and uncertainty calibration
    \STATE Sort the prompts and form contiguous candidate steps
    \STATE Set response-length limits and validate all steps with $\mathcal{P}$
    \FOR{each rejected step from last to first}
      \STATE Search bounded swaps with the preceding step
      \STATE Recompute limits and rerun $\mathcal{P}$ on affected steps
      \STATE Cache the first valid repaired plans
    \ENDFOR
    \RETURN $\Pi$
  \end{algorithmic}
\end{algorithm}

\section{Offline Planner for Each Rollout Step}
\label{sec:floorplanning}

This section defines $\mathcal{P}(B_t,C_t)$ through panel 2b's memory checks
and panel 2c's release planning. For a fixed batch and response limit,
the solver jointly chooses prompt pairs, an EP group-size sequence, completion
thresholds, and physical rank placement. A group size is useful only if a
memory-safe assignment also completes enough ranks early for release.

\subsection{Step-Level Memory Feasibility}
\label{sec:floorplanning:graph}

To check whether a candidate prompt pair $e=(i,j)$ fits on a rank, AdaFloor
estimates how much KV cache its responses occupy at the same time. Let
$L_i$ and $L_j$ contain the sampled response lengths from prior rollouts,
including repeated lengths. At decode position $\tau$, responses shorter
than $\tau$ have released their KV blocks. The estimated active-token peak is
\begin{equation}
A_e=\max_{\tau}
\left(
\tau\cdot\#\{\ell\in L_i\uplus L_j\mid\ell\ge\tau\}
\right),
\label{eq:active_peak}
\end{equation}
which distinguishes a pair with many overlapping long responses from one
with only a single long tail. Maximum response length alone cannot make
that distinction.

To allow for longer responses in the next rollout, AdaFloor scales each
historical length by a safety factor $\mu$ before applying the batch and
global output-length limits.
The resulting peak is $A_{\mu,e}$. For floor $f$ under runtime profile $m$,
the feasible-pair graph is $G_{t,f}=(V_t,E_{t,f})$, where $V_t=B_t$ and
\begin{equation}
E_{t,f}=\{e\mid A_{\mu,e}\le C_{f,m}^{\mathrm{eff}}\}.
\label{eq:kv_graph}
\end{equation}
Vertices are prompts and edges are memory-safe prompt pairs. A perfect
matching assigns every prompt exactly once. A larger EP group may admit
a complete matching even when a smaller group's tighter KV budget does not.

For each feasible pair, the predicted rank completion is
\begin{equation}
M_e=\max(\hat M_i,\hat M_j).
\label{eq:pair_completion}
\end{equation}
These completion estimates connect memory admission to release planning.
A KV-safe matching alone is insufficient if too few of its ranks finish
before the step tail. The solver must also satisfy completion quotas.

\subsection{Plan Rank Release}
\label{sec:floorplanning:ilp}

AdaFloor favors schedules that release more ranks earlier. A schedule starts
with $f_0=N$ ranks and shrinks to $f_s$ ranks at decode position $\tau_s$
for each transition $s=1,\ldots,S$. The $f_{s-1}-f_s$ ranks released at
that transition remain available until the predicted end of the step, $T_t$.
The total predicted released rank-time is
\begin{equation}
R=\sum_{s=1}^{S}(f_{s-1}-f_s)(T_t-\tau_s).
\label{eq:release_area}
\end{equation}
To maximize $R$, the outer search enumerates schedules for each sequence
of EP group sizes, using predicted pair completion positions $M_e$ as
candidate transition points. It checks schedules from highest to lowest
$R$, moving to the next one if the inner solver finds no feasible prompt
assignment. The first accepted schedule maximizes $R$ among the feasible
schedules for that sequence.

The inner solver keeps the candidate's transition positions
$\tau_1,\ldots,\tau_S$ fixed. It uses a binary integer linear program (ILP)
to find the lowest-cost assignment that meets all scheduled transitions.
For each KV-safe pair $e\in E_{t,f}$ at the final group size $f$, let $c_e$ be the cost
of placing the two prompts on the same rank. The binary variable $x_e$ is
one if the pair is selected and zero otherwise. The ILP is
\begin{align}
\min_{\boldsymbol{x}}\quad &\sum_{e\in E_{t,f}} c_e x_e,
\label{eq:matching_objective}\\
\text{subject to}\quad &\sum_{e\ni i}x_e = 1, && i\in V_t,
\label{eq:perfect_matching}\\
&\sum_{e\mid M_e\le\tau_s}x_e \ge N-f_s, && s=1,\ldots,S.
\label{eq:quota}
\end{align}
The first constraint assigns every prompt exactly once. The second requires
that the same assignment have at least $N-f_s$ ranks finished by each
$\tau_s$, including ranks released at earlier~transitions.

Among assignments that meet these constraints, the pair cost favors similar
predicted lengths within a pair, earlier completion, and lower peak KV use.
For $e=(i,j)$, the respective terms are
\begin{equation}
c_e=10^6\lvert\hat M_i-\hat M_j\rvert+10^3 M_e+A_{\mu,e}.
\label{eq:pair_cost}
\end{equation}

\subsection{Select EP Sizes and Place Prompts}
\label{sec:floorplanning:materialization}

AdaFloor compares the best schedule for each feasible final EP group size.
Let $\mathcal{F}_t^{\mathrm{safe}}$ be the set of final sizes with a feasible
plan for step $t$, and let $R_{t,f}^{\star}$ be the best predicted release
score for size $f$. AdaFloor chooses
\begin{equation}
\hat f_t=\arg\max_{f\in\mathcal{F}_t^{\mathrm{safe}}}R_{t,f}^{\star}.
\label{eq:cross_floor_selection}
\end{equation}
If two plans are predicted to release the same amount of rank-time,
AdaFloor chooses the one that keeps more ranks active at the end, so it
needs fewer shrink steps.

Before rollout, AdaFloor assigns prompt pairs expected to finish later to
ranks that the offline plan keeps active longer. It passes the
prompt placement, planned EP group sizes and communication groups, and
output-length limit to the executor (Figure~\ref{fig:adafloor_overview}).
Observed completions determine which ranks actually leave.

\section{Runtime Executor}
\label{sec:runtime}

The runtime follows the right column of Figure~\ref{fig:adafloor_overview}.
It applies the planned length limit, monitors local completion, checks whether
a smaller group is safe, switches groups, and makes released accelerators
available for other tasks. It brings all ranks back before reward computation
and policy updates.

\subsection{Prepare Experts for Smaller Groups}
\label{sec:runtime:redundancy}

When the EP group shrinks, the remaining ranks still need access to every
expert. AdaFloor keeps expert copies on the accelerators so the smaller
group serves tokens without fetching weights for each token. The memory
budget counts all reserved expert slots, including slots the current group
does not use but a smaller group may need. After a policy update, AdaFloor
refreshes the copies before the next rollout.
Section~\ref{sec:eval:redundancy} compares this choice with host-memory and
remote weight retrieval.

In \emph{Natural} mode, AdaFloor sets up resources when the group shrinks.
This leaves more memory for the KV cache but adds setup work at the
transition. \emph{Planned residency} reserves
memory in advance for communication and token routing, reducing setup work
at the transition. The offline planner uses the KV budget for the chosen
mode. In both modes, observed completions determine which ranks stay active.

\subsection{Apply Limits and Monitor Completion}
\label{sec:runtime:completion}

At the start of each rollout step, the executor applies the output-length
limit from the offline plan to every response in the batch. A rank finishes
locally once all its assigned responses are complete. The executor considers
shrinking to a planned group of $f$ ranks only when at most $f$ ranks still
have unfinished responses.

Before switching, the executor checks that the proposed EP group is supported,
that the remaining ranks can serve every expert, and that their KV cache has
space for the unfinished~responses.

If too few KV blocks are available during generation, AdaFloor reuses
vLLM's request preemption mechanism~\cite{kwon2023pagedattention}. The
scheduler pauses a request and frees its KV blocks. When the request resumes,
the executor rebuilds its cache from the existing tokens and continues
generation. Although this fallback remains available, AdaFloor's KV-aware
prompt placement and group-size selection avoided KV preemption across its
variants in the primary evaluation (Section~\ref{sec:eval:floors}).

\subsection{EP Group Changes and Decode Graphs}
\label{sec:runtime:graphs}

When the EP group shrinks, only ranks still serving rollouts join the
smaller data-parallel (DP) and EP groups. Released ranks no longer take
part in rollout collectives. Any auxiliary task on those ranks uses its
own communication groups if needed. The rollout processes remain alive
so they can rejoin the original group later.

The fixed-group verl baseline supports decode graphs for an unchanged EP
group, but not the group changes required by AdaFloor. A graph refers to the communication
groups, expert placement, and KV buffers in place at capture time. AdaFloor
retires obsolete graphs when the group or its buffers change, then captures
new graphs after setting up the current expert state, groups, and KV storage.
It frees graph memory separately from the expert and attention state needed
for subsequent execution. The graphs cover the full decode pass, including
attention and MoE, while prefill and host-side scheduling run outside them.

AdaFloor also adapts graph sizes to each rollout step. A graph size is the
number of sequences it processes in one decode iteration. AdaFloor uses
historical response lengths and the planned prompt assignment to estimate
how often each batch size occurs. For a given graph count, it uses dynamic
programming to select the sizes that minimize predicted padding. During
decoding, the executor replays the smallest graph that fits the live batch.
More sizes reduce padding but add capture time and memory costs, so we profile
the graph count separately. Section~\ref{sec:eval:overhead} reports the
selection trade-off and compares the configured graph and eager paths.

Departing ranks follow the same group-creation sequence without joining
the smaller groups. Before the original group is rebuilt, ranks that left
earlier complete any setup steps they missed.

\subsection{Resource Reuse and Safe Return}
\label{sec:runtime:reuse}

When a rank leaves the rollout group, its accelerator stops decoding, but
its KV cache may still occupy device memory. The executor resizes or releases
that cache as needed. It starts another task only on released accelerators with
enough free memory and enough expected time to start, do useful work, and
shut down before the accelerators must rejoin the rollout group
(Figure~\ref{fig:adafloor_overview}).

Before the accelerators rejoin the original rollout group, the executor asks
the other tasks to stop and waits for confirmation that they have saved their
completed results and released their resources. Reward computation and policy
updates begin only after the full group is ready. Section~\ref{sec:eval:setup}
explains how we measure the time for which each rank is released.

\section{Evaluation}
\label{sec:evaluation}

We first compare rollout throughput and rank-time resource efficiency on a
common graph backend. We then examine how AdaFloor shortens rollouts and
releases accelerators early, whether it does so safely and at what cost, and
what work the released accelerators perform. Finally, we test how the gains
change with the workload and batch settings.

\begin{figure*}[t]
 \centering
 \setlength{\abovecaptionskip}{2pt}
 \includegraphics[width=\textwidth,trim=0 5bp 0 0,clip]{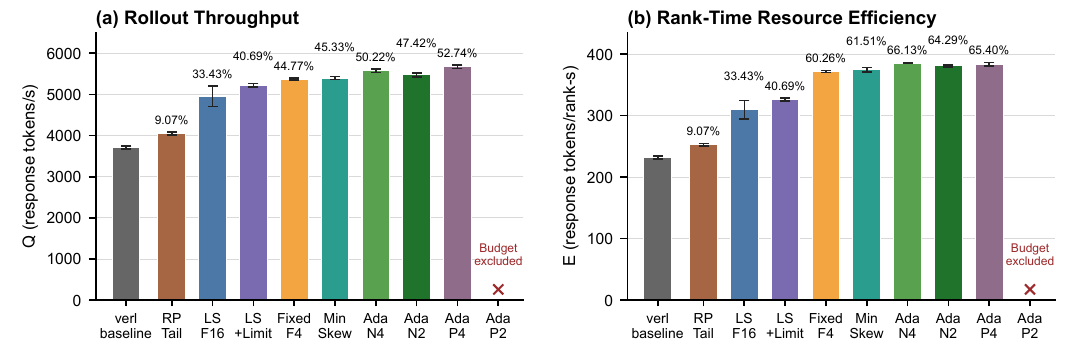}\par
 \caption{Rollout performance on DeepScaleR.
 (a) Rollout throughput $Q$. (b) Occupied rank-time resource efficiency $E$.
 Bars show seed means with sample standard deviations. Percentage labels
 give improvements over the verl baseline. LS is LengthSort, LS+Limit is LengthSort
 with batch limits, and RP is RollPacker tail batching. Ada-P2 exceeds the memory budget.}
 \label{fig:e2e}
\end{figure*}

\subsection{Experimental Setup}
\label{sec:eval:setup}

\paragraph{Platform and implementation.}
AdaFloor integrates verl with vLLM 0.14.1 and its Ascend backend, PyTorch
support for neural processing units (NPUs), Compute Architecture for Neural
Networks (CANN), and the Huawei
Collective Communication Library (HCCL). The planner uses SciPy
mixed-integer linear programming (MILP) and NetworkX
matching~\cite{virtanen2020scipy,hagberg2008networkx}.
Our node has eight dual-die Ascend 910C packages, exposing 16 devices with
64~GiB HBM each and Huawei Cache Coherence System (HCCS) links.
Qwen3-30B-A3B uses bfloat16 (BF16), 128 experts, and top-8 routing, with
tensor parallelism (TP) and pipeline parallelism (PP) both set to 1,
DP16, and EP16.

\paragraph{Execution configuration.}
All primary runs use \texttt{FULL\_DECODE\_ONLY} with task-queue setting 1 (TQ1),
eight graph capture sizes per step, and native sleep level 2 with KV-first wake
(Section~\ref{sec:runtime:graphs}). Table~\ref{tab:effective_kv_caps} lists
AdaFloor's KV budgets. From a shared epoch~0 checkpoint and history, we run
two GRPO epochs on DeepScaleR for each rollout seed (0, 2, 3), with a
$10^{-6}$ learning rate, 160 prompts over five steps per epoch, and 16 retained
responses per prompt. Input and output limits are 1024 and 16384 tokens.
Sampling uses temperature 0.9, top-$p$ 0.9, and top-$k$ 50. The planner uses
$\alpha=0.3$, $q=0.95$, $K=3$, and $C_{\min}=4096$ and rounds response limits
up in units of 512 tokens (Section~\ref{sec:epoch}). The KV safety factor is
$\mu=1.16$ (Section~\ref{sec:floorplanning:graph}).

The verl baseline keeps the original prompt order at EP16. Our same-backend
RollPacker tail batching~\cite{gao2026rollpacker} keeps TP1/EP16 fixed.
In each of four short rounds, it samples 20 candidates for each of 40 prompts
and keeps 16 responses from each of the first 32 prompts to reach that quota.
The last round resamples all 32 deferred prompts. LengthSort groups prompts
by predicted length at EP16. LengthSort with batch limits (LS+Limit) also caps
response length per batch
(Section~\ref{sec:epoch}). Fixed-F4 targets EP4, and MinSkew minimizes
within-pair length differences. Both use 288,256 KV tokens per rank at EP8.
Ada-N2 and Ada-N4 use Natural residency with minimum EP sizes of two and four.
Ada-P4 uses Planned residency with a minimum of four.

\begin{table}[t]
 \centering
 \caption{AdaFloor's per-rank KV cache capacity in tokens. Planned-F2 is excluded
 because reserving memory for its experts leaves no usable KV capacity.}
 \label{tab:effective_kv_caps}
 \footnotesize
 \setlength{\tabcolsep}{4pt}
 \begin{tabular}{lrr}
 \toprule
 EP Size & Natural & Planned \\
 \midrule
 2 & 131,072 & Excluded \\
 4 & 280,576 & 133,120 \\
 8 & 349,952 & 235,264 \\
 16 & 380,800 & 380,800 \\
 \bottomrule
 \end{tabular}
\end{table}

\paragraph{Metrics and timing.}
We define throughput $Q$ and resource efficiency $E$ using saved response
tokens $W$, rollout time $T$, and occupied rank-time $H$.
\begin{equation}
 Q=W/T,\qquad E=W/H.
 \label{eq:v014_metrics}
\end{equation}
$T$ includes rollout preparation, generation, and cleanup, but excludes
actor updates, checkpoint serialization, and offline planning. $W$ excludes
prompt tokens and a proven padding end-of-sequence (EOS) token beyond a short
cap. For RollPacker, $T$ includes discarded candidates' generation time,
while $W$ counts only retained responses.
Each rank is charged until its shrink remote procedure call (RPC) and local
sleep cleanup complete. We count its released rank-time from then until the
first restore or wake, so $H$ includes quorum waiting and restoration reoccupation.
Ranks that never detach are charged for the full rollout.

\subsection{Overall Rollout Performance}
\label{sec:eval:e2e}

On the same graph backend, Ada-P4 delivers the highest mean rollout
throughput, while Ada-N4 achieves the highest mean rank-time resource
efficiency. Mean rollout reward varies little across methods.

In Figure~\ref{fig:e2e}, Ada-P4 reaches $5668.7\pm 42.8$ response tokens/s
versus $3711.3\pm 33.0$ for the verl baseline, a \ThroughputGain\% gain.
The same configuration improves rank-time resource efficiency by approximately
\ResourceEfficiencyGain\%.
Its efficiency is about 383.6 response tokens/rank-s versus 232.0.
Ada-N4 improves resource efficiency by 66.13\% over the verl baseline.
Mean reward across methods ranges from 0.8332 to 0.8366.

RollPacker tail batching beats the verl baseline in all three seeds, reaching
4047.8 tokens/s ($+9.07\%$) and taking 3633.0~s per epoch ($12.51\%$ less).
Ada-P4 improves on this control by 40.04\% in throughput and
51.5\% to 51.8\% in resource efficiency (383.27--384.03 versus 252.99
tokens per occupied rank-second). Mean rollout reward is 83.46\% for Ada-P4
and 83.42\% for RollPacker.

Ada-P4 reduces mean rollout time from 4152.2 to 2710.1~s. Its occupied
rank-time decreases from 66,435.5 to approximately 40,003 to 40,083 rank-s
per epoch, including restoration. Occupied cost per generated response token
falls by about \ResourceCostReduction\%.

\paragraph{Cold start and composition.}
\label{sec:eval:cold-start}

RollPacker tail batching improves first-epoch rollout throughput by 11.71\%
over verl without requiring response-length history. Once such history is
available, Ada-P4 outperforms tail batching in both subsequent epochs
(Table~\ref{tab:cold-start}). To combine these complementary gains, we
introduce an epoch-wise hybrid that uses tail batching in epoch~0, then
switches to Ada-P4 using the resulting policy checkpoint and response-length
history. Across all three epochs, the hybrid achieves 8.67\% higher rollout
throughput than Ada-P4 alone.

\begin{table}[t]
 \centering
 \caption{Rollout throughput by epoch (response tokens/s).}
 \label{tab:cold-start}
 \footnotesize
 \setlength{\tabcolsep}{4pt}
 \begin{tabular}{lrrr}
 \toprule
 Method & Epoch 0 & Epoch 1 & Epoch 2 \\
 \midrule
 verl baseline & 3,653 & 3,657 & 3,648 \\
 RollPacker tail batching & \textbf{4,081} & 4,234 & 4,122 \\
 Ada-P4 & 3,653 & 5,739 & 5,503 \\
 RollPacker $\to$ Ada-P4 & \textbf{4,081} & \textbf{6,158} & \textbf{5,731} \\
 \bottomrule
 \end{tabular}
\end{table}

\subsection{Where the Gains Come From}
\label{sec:eval:breakdown}

Batch construction accounts for most of the rollout throughput gain.
Elastic EP improves resource efficiency further by releasing locally
finished ranks. We use the fixed-group controls to distinguish these effects,
then examine how group-size selection and prompt pairing bring releases
forward.

\paragraph{Batch construction and response limits}
Sorting prompts and setting batch limits improve throughput even at a fixed
EP group size.
At EP16, LengthSort increases throughput from 3711.3 to 4952.2 response
tokens/s, and LS+Limit reaches 5221.3. Their efficiency gains equal their
throughput gains because $H=16T$.
Against LS+Limit, Ada-P4 adds \ThroughputGainLS\% throughput and approximately
\EfficiencyGainLS\% efficiency. Its paired seed-level rollout-time difference
is 232.3~s, with a 95\% Student-$t$ interval of 123.9 to 340.6~s.

Batch limits also help within Ada-P4. They improve mean throughput by
\TGTPSGain\% and mean efficiency by approximately \TGResourceGain\%
(Table~\ref{tab:v014_tg}). The comparison retains EOS stopping and the global
16384-token limit in both runs, with identical prompt assignments in epoch 1.
Across the verl baseline, LengthSort, and LS+Limit, mean response length
stays near 6000 tokens and mean rollout reward varies little.

\begin{table}[t]
 \centering
 \caption{Ada-P4 with and without batch limits.}
 \label{tab:v014_tg}
 \footnotesize
 \setlength{\tabcolsep}{4pt}
 \begin{tabular}{lrr}
\toprule
& With batch limits & Without batch limits \\
\midrule
Throughput (tok/s) & 5668.7 $\pm$ 42.8 & 5469.2 $\pm$ 122.4 \\
Efficiency (tok/rank-s) & [383.2, 384.1] & [376.9, 377.7] \\
Reward & 0.8346 $\pm$ 0.0028 & 0.8333 $\pm$ 0.0020 \\
Mean length & 6000.7 $\pm$ 11.6 & 5989.5 $\pm$ 15.2 \\
\bottomrule
\end{tabular}

\end{table}

\begin{figure}[!t]
 \centering
 \setlength{\abovecaptionskip}{3pt}
 \includegraphics[width=\columnwidth]{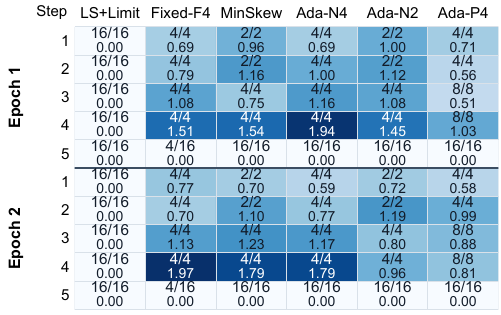}
 \caption{Planned/observed minimum EP sizes above released rank-time
 (in thousands of rank-seconds) for seed 0.}
 \label{fig:ablation}
\end{figure}

\paragraph{EP group sizes and rank release}
Shortening a rollout does not by itself free ranks before it ends. With
elastic EP, the smallest group reached varies by step. In the seed 0 runs,
the AdaFloor variants release ranks in the earlier steps of each epoch but
retain the full group in the last step.
Figure~\ref{fig:ablation} compares planned and observed minimum EP group
sizes with measured released rank-time.

\paragraph{Quota-aware prompt pairing}
Alongside choosing a group size, the planner must arrange for enough ranks
to finish early to reach it. Quota-aware pairing reduces waiting before
rank release. Following Insight~3 (Section~\ref{sec:char:kv}), it pairs prompts
so their overlapping responses fit in the KV cache and enough ranks finish
early for the EP group to shrink.
We take the original prompts and recorded response lengths from 120 completed
rollout steps. Quota-aware pairing and MinSkew each form new prompt pairs
from these records without generating new responses.
From the resulting completion positions, we estimate how many ranks each plan
would release and for how many decoding steps (Equation~\ref{eq:release_area}).
Quota-aware pairing gives a higher release estimate than MinSkew in eight
steps and matches it in the other 112. Across the 96 steps with positive
estimated release under MinSkew, the gain averages 0.077\% and reaches 2.972\%.

In a separate experiment on NPUs, both methods generate the same number of
tokens. With quota-aware pairing, the eighth rank finishes about 35~s earlier
than with MinSkew. Ranks spend \QuotaWaitReduction\% less total time waiting
after local completion, while throughput and resource efficiency increase by
\QuotaQGain\% and \QuotaEGain\%, respectively.

\paragraph{Sensitivity to length prediction deviations}
Both batch construction and rank release depend on predicted lengths.
Underprediction raises measured throughput at the cost of lower rollout
reward and more KV preemptions. Overprediction reduces resource efficiency
by releasing fewer ranks or releasing them later. Even without deliberately
changing the forecasts, sorting by predicted rather than observed lengths
increases the sum of observed batch maxima by 14.93\% to 32.30\% across
six epochs.

All runs use the same frozen epoch 0 checkpoint and the same 160 prompts over
five steps with seed 0. We scale the predicted lengths and response profiles
used for KV admission by $1.0$, $0.7$, or $1.3$. The mixed case assigns
either perturbation to each prompt using a fixed hash. We hold the
response-length multiplier at 1.2 and recompute prompt placement, EP group
sizes, and batch limits for each setting.
All settings finish without out-of-memory (OOM) errors or aborted requests
(Table~\ref{tab:forecast_sensitivity}).

\begin{table}[t]
 \centering
 \caption{V0.14 frozen-policy forecast sensitivity. $Q$ is tokens/s and
 $E$ is tokens/rank-s. L counts hits of shorter batch limits,
 P counts recovered KV preemptions, and G counts hits of the global~limit.}
 \label{tab:forecast_sensitivity}
 \footnotesize
 \setlength{\tabcolsep}{2pt}
 \begin{tabular}{lrrrrr}
 \toprule
 Forecast & $Q$ & $E$ & Reward & L/P & G \\
 \midrule
 $\times1.0$ & 5403.54 & $[380.3,381.2]$ & 0.8371 & 2/0 & 102 \\
 $\times0.7$ & 6142.27 & $[392.1,392.4]$ & 0.8281 & 267/75 & 0 \\
 $\times1.3$ & 5447.30 & $[376.8,377.7]$ & 0.8352 & 0/0 & 95 \\
 Mixed & 5371.87 & $[357.4,357.9]$ & 0.8305 & 92/0 & 89 \\
 \bottomrule
 \end{tabular}
\end{table}

Relative to the unscaled forecast, underprediction increases measured
throughput but triggers 75 KV preemptions and lowers mean rollout reward by
0.90 percentage points.
Overprediction leaves throughput nearly unchanged but reduces resource
efficiency by delaying rank release. In step 4, for example, the first
confirmed rank release occurs about 35~s~later.

\paragraph{Offline planning time}
Offline planning for five rollout steps takes 66.76~s, including 19.33~s to
build candidate schedules and pair graphs
and 38.76~s to solve the placement problems. Batch repair takes less than 1~ms.

\subsection{Safe and Efficient Elastic Execution}
\label{sec:eval:floors}
\label{sec:eval:overhead}
\label{sec:eval:redundancy}

AdaFloor stays within the available memory in the primary runs and preserves
outputs in matched decoding tests.

\paragraph{Expert availability}
To keep every expert available after the EP group shrinks, we implement and
evaluate five schemes. Resident replication keeps redundant expert copies on
the active NPUs. Remote NPU fetching retrieves weights from ranks that have
left the group. Dual-source fetching splits transfers between remote NPUs
and central processing unit (CPU) memory. CPU double buffering loads the next
layer's weights while the current layer computes, whereas CPU swapping loads
weights from host memory as needed. All five methods use a frozen policy,
greedy decoding without EOS stopping, FULL-decode graphs of size 32, and
133,120 KV tokens per rank. Ranks supplying remote weights count toward
occupied rank-time $H$. Resident replication delivers the highest rollout
throughput and rank-time resource efficiency
(Table~\ref{tab:expert_availability_modes}), so we adopt it in the AdaFloor
runtime (Section~\ref{sec:runtime:redundancy}).

\begin{table}[t]
 \centering
 \caption{Length-controlled expert availability comparison. Each run uses
 the same 32 prompts, 16 responses per prompt, and 262,144 output tokens
 through EP16$\rightarrow$EP8$\rightarrow$EP4. Values are means of three
 single-step runs, sorted by throughput.}
 \label{tab:expert_availability_modes}
 \footnotesize
 \setlength{\tabcolsep}{5pt}
 \begin{tabular}{lrr}
 \toprule
 Mechanism & $Q$ & $E$ \\
 \midrule
 Resident replication & \textbf{2,384.39} & \textbf{217.05} \\
 Remote NPU fetching & 1,520.02 & 95.00 \\
 CPU + remote dual source & 1,364.55 & 85.28 \\
 CPU double buffering & 1,008.65 & 136.89 \\
 CPU swapping & 749.65 & 108.91 \\
 \bottomrule
 \end{tabular}
\end{table}

\paragraph{Memory feasibility and reclamation}
The primary AdaFloor runs finish without KV preemption. In separate
instrumented runs with a fixed policy, peak KV occupancies reach 93.9\% for
Ada-N2 and 98.4\% for Ada-P4, with at least 7,936 and 3,712 KV token slots
free per rank, respectively.
In the first five consecutive policy updates, neither Natural-F2 nor
Planned-F4 triggers KV preemption, runs out of memory, or aborts responses.
From just before local rollout release to after cache cleanup, median free
device memory increases by
42.98~GiB for Natural-F2 and 32.09~GiB for Planned-F4.

\paragraph{Output consistency and answer accuracy}
Elastic FULL-decode matches the EP16 reference token for token on all 1,536
responses per configuration over three matched greedy steps with a fixed
policy. Natural-F2 exercises EP sizes of 2, 4, 8, and 16, while Planned-F4
exercises 8 and 16. All 496 graph-to-eager comparisons sampled at transitions
also match across the first five updates of both runs. In greedy evaluation
after five updates, the eager verl baseline, eager Ada-P4, and graph-enabled
Ada-N2 with additional inference answer 211, 208, and 212 of the same 256
held-out DeepScaleR questions correctly, respectively.

\paragraph{Configured eager and graph controls}
Compared with eager execution, FULL-decode graphs improve rollout throughput
by \GraphBaselineGain\% for the verl baseline and \GraphPfourGain\% for
Ada-P4 (Table~\ref{tab:v014_graph}).

\begin{table}[t]
 \centering
 \caption{Eager and graph rollout throughput (tokens/s).}
 \label{tab:v014_graph}
 \footnotesize
 \setlength{\tabcolsep}{4pt}
 \begin{tabular}{lrrr}
\toprule
Method & Eager & Graph & Gain \\
\midrule
verl baseline & 1964.13 & 3746.71 & +90.76\% \\
Ada-P4 & 2515.24 & 5656.35 & +124.88\% \\
\bottomrule
\end{tabular}

\end{table}

\paragraph{Graph count and capture sizes}
In a fixed-token rollout step profiling 1 to 32 graphs, eight graphs nearly
match the fastest ten-graph configuration (3363.6 vs. 3364.0 tokens/s) while
using less memory per rank (1.47 vs. 1.81~GiB). We use eight graphs per step
and let AdaFloor select their sizes, such as $[1,4,9,14,16,20,23,32]$ in this
profile.

\paragraph{EP transition and rollout preparation costs}
Shrinking takes longer than restoring the full group, and rebuilding
communication groups is the largest component of both
(Table~\ref{tab:v014_transitions}). The primary Planned-F4 runs provide
36 shrink events and 24 restorations.
Preparation before rollout generation, measured separately from EP shrink
time, takes 20.9~s for the verl baseline, 27.5--36.6~s for Ada-N2, and
19.1--29.3~s for Ada-P4 in the instrumented runs.

\begin{table}[t]
 \centering
 \caption{EP transition times in seconds. Each entry is the median across
 events of the slowest participating rank.}
 \label{tab:v014_transitions}
 \footnotesize
 \begin{tabular}{lrr}
 \toprule
 Component & Shrink & Restore \\
 \midrule
 Group rebuild & 8.70 & 2.90 \\
 Expert-state refresh & 0.56 & 1.35 \\
 Graph retirement & 2.44 & 1.89 \\
 Recapture & 5.02 & -- \\
 Worker transition total & 19.37 & 6.39 \\
 \bottomrule
 \end{tabular}
\end{table}

\subsection{Using the Released Accelerators}
\label{sec:eval:reuse}

Released accelerators complete inference requests and rejoin the primary
rollout group before policy updates. In separate Ada-N4 runs, each of the
three models serves GSM8K~\cite{cobbe2021gsm8k} in its best-performing
eight-device configuration on our platform. Qwen2.5-1.5B uses eight
independent TP1 replicas, Qwen3-8B uses a single DP2$\times$TP4 engine, and
PanGu-MoE~\cite{tang2025pangu} uses DP8$\times$TP1 with EP8.
Table~\ref{tab:sidecar_reuse} reports completed requests and answer accuracy.
Total primary rollout time stayed within 2.6\% of the matched run without
auxiliary inference.

\begin{table}[t]
 \centering
 \caption{Completed inference requests and answer accuracy during Ada-N4
 post-training. Each model uses a separate ten-step run.}
 \label{tab:sidecar_reuse}
 \footnotesize
 \setlength{\tabcolsep}{3pt}
 \begin{tabular}{lrrr}
\toprule
Model & Params (B) & Requests & Accuracy \\
\midrule
Qwen2.5-1.5B & 1.54 & 28,730 & 75.30\% \\
Qwen3-8B & 8.2 & 218 & 96.79\% \\
PanGu-MoE & 72 & 168 & 90.48\% \\
\bottomrule
\end{tabular}

\end{table}

\paragraph{Composition with worker-level reuse}
\label{sec:eval:worker_composition}
Adding AdaFloor's rank-level release to TLT-style worker-level
reuse~\cite{hu2026tlt} yields 4.27 times as much reusable rank-time as
worker-level reuse alone. This comparison measures available rank-time,
whereas the inference runs measure completed work. Across five rollout steps
with two EP8 workers, worker-level reuse exposes 1013.3 rank-seconds after
an entire worker becomes available. AdaFloor adds 3315.0 rank-seconds by
releasing ranks while their workers are still generating responses.

\subsection{Effects of Workload and Batch Settings}
\label{sec:eval:workloads}

AdaFloor improves throughput and resource efficiency on code as well as
mathematics. On the mathematics workload, the gains are smaller with a
shorter response limit.

\paragraph{Code-generation workload}
On Eurus-2-RL~\cite{cui2025prime}, Ada-N2 improves rollout throughput by
\CodeQGain\% and resource efficiency by approximately \CodeEGain\% over
the verl baseline, with similar sampled pass@1 (70/160 versus 71/160).
The frozen-policy comparison covers 160 problems over five rollout steps
with 16 responses per problem.

\paragraph{Workload sensitivity}
On the mathematics workload, Ada-N2 improves both throughput and resource
efficiency at every tested setting. The smallest gains occur under the
8192-token response limit (Table~\ref{tab:sensitivity}), where the baseline
already exceeds 6100 tokens/s.
We vary the number of responses per prompt, the response limit, and the
number of prompts per step one at a time. Each five-step pair starts from
the same frozen checkpoint and processes the same prompts in both methods.
The 16-prompt setting assigns one prompt to each of the 16 ranks.
Mean reward differs by at most
0.24 percentage points within any pair, and no run triggers KV preemption.

\begin{table}[t]
 \centering
 \caption{Five-step frozen-policy workload pairs. Throughput $Q$ is in
 tokens/s. Gains are relative to the verl baseline. Bounds on $\Delta E$
 reflect timestamp resolution.}
 \label{tab:sensitivity}
 \footnotesize
 \setlength{\tabcolsep}{3pt}
 \begin{tabular}{lrrrr}
 \toprule
 Setting & verl $Q$ & Ada-N2 $Q$ & $\Delta Q$ (\%) & $\Delta E$ (\%) \\
 \midrule
 Default & 3682 & 5404 & 46.7 & 65.22--65.66 \\
 8 responses/prompt & 2270 & 3226 & 42.1 & 65.20--65.85 \\
 8192-token limit & 6155 & 6626 & 7.7 & 17.29--17.58 \\
 16 prompts/step & 2205 & 3282 & 48.9 & 68.11--68.63 \\
 \bottomrule
 \end{tabular}
\end{table}

\section{Discussion}
\label{sec:discussion}

\paragraph{Choosing a configuration}
When rollout latency is the priority, Planned residency reduces setup work
when the EP group changes. When the priority is to release more
accelerators, Natural leaves more memory for the KV cache, allowing smaller EP
groups when enough ranks finish early. Beyond the inference tasks demonstrated
in Section~\ref{sec:eval:reuse}, released accelerators could support reward
scoring, policy evaluation, or training draft models for speculative decoding
if the release window is long enough.

\paragraph{Scaling beyond one node}
When one model replica fits on a node, a multi-node deployment could keep EP
within each node and use DP across nodes. For EP groups that span nodes,
the planner would need to account for network topology when choosing which
ranks stay active and use measured memory budgets and transition costs for
those groups. The runtime would also need to manage cross-node communication
groups while preserving expert availability and collective ordering.

\section{Related Work}
\label{sec:related}

\subsection{MoE and Inference Execution}
MoE runtimes optimize expert placement and execution~\cite{hwang2023tutel,
he2022fastermoe,zhai2023smartmoe,nie2023flexmoe,zeng2025efficientmoe}.
Serving systems improve expert execution, batching, and decoding
~\cite{hwang2024pregated,kamahori2025fiddler,fang2025klotski,agrawal2024sarathi,
zhong2024distserve,zheng2024sglang,leviathan2023speculative,li2024eagle}.
LAER-MoE redistributes expert shards to balance training
load~\cite{liu2026laermoe}, and StreamEP
removes global communication barriers to reduce decoding stalls
~\cite{liang2026streamep}. AdaFloor instead changes EP group membership during
generation to free devices while keeping experts available.

\subsection{RL Post-Training and Rollouts}
Prior systems improve RL post-training through resource reallocation, stage
overlap, and asynchronous coordination~\cite{mei2025real,yu2026rlinf,
wang2026dynarl,zhong2025rlhfuse,gao2026rollart}. Rollout optimizations speed
up generation through request scheduling and speculative
decoding~\cite{qin2026seer,shao2026das} or reduce idle time across workers
through batching, migration, and load balancing~\cite{gao2026rollpacker,
he2026rhymerl}. Rather than balancing work across workers, AdaFloor releases
locally finished ranks within a worker while its EP group remains active.

\subsection{Elastic Resource Management}
Accelerator schedulers resize or colocate jobs~\cite{qiao2021pollux,
miao2024spotserve,strati2024orion,wang2025sirius}. KV cache systems improve
memory allocation and data movement~\cite{kwon2023pagedattention,
prabhu2025vattention,qin2025mooncake}.
TLT trains its drafter on idle rollout workers, and BubbleSpec generates
drafts for later steps on idle data-parallel ranks~\cite{hu2026tlt,xu2026bubblespec}.
Weave reuses resources left idle by stage dependencies across RL jobs
~\cite{wu2026weave}, while RLBoost borrows preemptible capacity for rollouts
~\cite{wu2026rlboost}. AdaFloor makes released devices available to other
tasks and restores the EP group before the next policy update.

\section{Conclusion}
\label{sec:conclusion}

We present AdaFloor, a system that shortens synchronous MoE rollouts and
frees accelerators from active EP groups. Its offline planner
uses historical rollout lengths to form batches, set output-length limits, and jointly
plan prompt placement and EP group sizes within device memory budgets. Its
graph-enabled runtime executor releases locally finished ranks while preserving
expert availability and collective ordering, then restores the full group
before policy updates. On DeepScaleR, AdaFloor improves rollout throughput by
\ThroughputGain\% and rank-time resource efficiency by approximately
\ResourceEfficiencyGain\% over the verl baseline, with little change in rollout
reward.
RollPacker improves throughput without length history, while AdaFloor delivers
larger gains once that history is available.
Inference on released devices shows that synchronous post-training
and serving can share the same accelerators.

\label{osdi:bodyend}
\bibliographystyle{plain}
\bibliography{refs}

@inproceedings{liang2026streamep,
  author       = {Yizhuo Liang and Shaoyu Wang and Jaeyong Song and Yanqi Zhou and
      Geon-Woo Kim and Guangrong He and Seo Jin Park},
  title        = {{StreamEP: Straggler-Tolerant MoE Decoding without Communication Barriers}},
  booktitle    = {ACM Symposium on Operating Systems Principles},
  pages        = {600--619},
  year         = {2026},
  doi          = {10.1145/3830418.3843869},
  url          = {https://doi.org/10.1145/3830418.3843869}
}

@inproceedings{liu2026laermoe,
  author       = {Xinyi Liu and Yujie Wang and Fangcheng Fu and Xuefeng Xiao and
      Huixia Li and Jiashi Li and Bin Cui},
  title        = {{LAER-MoE: Load-Adaptive Expert Re-layout for Efficient Mixture-of-Experts Training}},
  booktitle    = {ACM International Conference on Architectural Support for Programming Languages and Operating Systems},
  pages        = {1055--1072},
  year         = {2026},
  doi          = {10.1145/3779212.3790180},
  url          = {https://doi.org/10.1145/3779212.3790180}
}

@inproceedings{qin2026seer,
  author       = {Ruoyu Qin and Weiran He and Weixiao Huang and Yangkun Zhang and
      Yikai Zhao and Bo Pang and Xinran Xu and Yingdi Shan and Yongwei Wu and Mingxing Zhang},
  title        = {{Seer: Online Context Learning for Fast Synchronous LLM Reinforcement Learning}},
  booktitle    = {USENIX Symposium on Operating Systems Design and Implementation},
  pages        = {883--901},
  year         = {2026},
  url          = {https://www.usenix.org/conference/osdi26/presentation/qin}
}

@inproceedings{yu2026rlinf,
  author       = {Chao Yu and Yuanqing Wang and Zhen Guo and Hao Lin and Si Xu and
      Hongzhi Zang and Quanlu Zhang and Yongji Wu and Chunyang Zhu and Junhao Hu and
      Zixiao Huang and Mingjie Wei and Yuqing Xie and Ke Yang and Bo Dai and
      Zhexuan Xu and Jiakun Du and Xiangyuan Wang and Xu Fu and Letong Shi and
      Zhihao Liu and Kang Chen and Weilin Liu and Gang Liu and Boxun Li and
      Jianlei Yang and Zhi Yang and Guohao Dai and Yu Wang},
  title        = {{RLinf: Flexible and Efficient Large-Scale Reinforcement Learning via Macro-to-Micro Flow Transformation}},
  booktitle    = {USENIX Symposium on Operating Systems Design and Implementation},
  pages        = {829--846},
  year         = {2026},
  url          = {https://www.usenix.org/conference/osdi26/presentation/yu-chao}
}

@inproceedings{wang2026dynarl,
  author       = {Yuanqing Wang and Hao Lin and Junhao Hu and Chunyang Zhu and Quanlu Zhang and
      Zhen Guo and Yuchen Zhang and Xu Fu and Si Xu and Bo Dai and Zixiao Huang and
      Chao Yu and Boxun Li and Guohao Dai and Zhi Yang and Yu Wang},
  title        = {{DynaRL: Flexible and Dynamic Scheduling of Large-Scale Reinforcement Learning Training}},
  booktitle    = {USENIX Symposium on Operating Systems Design and Implementation},
  pages        = {847--862},
  year         = {2026},
  url          = {https://www.usenix.org/conference/osdi26/presentation/wang-yuanqing}
}

@inproceedings{gao2026rollart,
  author       = {Wei Gao and Yuheng Zhao and Tianyuan Wu and Shaopan Xiong and Weixun Wang and
      Dakai An and Lunxi Cao and Dilxat Muhtar and Zichen Liu and Haizhou Zhao and Ju Huang and
      Siran Yang and Yongbin Li and Wenbo Su and Jiamang Wang and Lin Qu and Bo Zheng and Wei Wang},
  title        = {{RollArt: Disaggregated Multi-Task Agentic RL Training at Scale}},
  booktitle    = {USENIX Symposium on Operating Systems Design and Implementation},
  pages        = {863--881},
  year         = {2026},
  url          = {https://www.usenix.org/conference/osdi26/presentation/gao}
}

@inproceedings{wu2026weave,
  author       = {Tianyuan Wu and Lunxi Cao and Yining Wei and Wei Gao and Yuheng Zhao and
      Dakai An and Shaopan Xiong and Zhiqiang Lv and Ju Huang and Siran Yang and Yinghao Yu and
      Jiamang Wang and Lin Qu and Wei Wang},
  title        = {{Weave: Efficient Co-Scheduling for Disaggregated RL Post-Training}},
  booktitle    = {USENIX Symposium on Operating Systems Design and Implementation},
  pages        = {809--827},
  year         = {2026},
  url          = {https://www.usenix.org/conference/osdi26/presentation/wu-tianyuan}
}

@inproceedings{xu2026bubblespec,
  author       = {Yuhang Xu and Kaibin Tian and Yang Tian and Zhice Yang and Yifeng Yu and
      Yan Li and Shengzhong Liu and Fan Wu and Guihai Chen},
  title        = {{BubbleSpec: Turning Long-Tail Bubbles into Speculative Rollout Drafts for Synchronous Reinforcement Learning}},
  booktitle    = {International Conference on Machine Learning},
  series       = {Proceedings of Machine Learning Research},
  volume       = {306},
  pages        = {141950--141964},
  year         = {2026},
  url          = {https://proceedings.mlr.press/v306/xu26bc.html}
}

@inproceedings{wu2026rlboost,
  author       = {Yongji Wu and Xueshen Liu and Haizhong Zheng and Juncheng Gu and Beidi Chen and
      Z. Morley Mao and Arvind Krishnamurthy and Ion Stoica},
  title        = {{RLBoost: Harvesting Preemptible Cloud Resources for Cost-Efficient Reinforcement Learning on LLMs}},
  booktitle    = {USENIX Symposium on Networked Systems Design and Implementation},
  pages        = {1323--1339},
  year         = {2026},
  url          = {https://www.usenix.org/conference/nsdi26/presentation/wu-yongji}
}

@inproceedings{shao2026das,
  author       = {Zelei Shao and Vikranth Srivatsa and Sanjana Srivastava and Qingyang Wu and
      Alpay Ariyak and Xiaoxia Wu and Ameen Patel and Jue Wang and Percy Liang and Tri Dao and
      Ce Zhang and Yiying Zhang and Ben Athiwaratkun and Chenfeng Xu and Junxiong Wang},
  title        = {{Beat the long tail: Distribution-Aware Speculative Decoding for RL Training}},
  booktitle    = {Proceedings of Machine Learning and Systems},
  volume       = {8},
  pages        = {1626--1638},
  year         = {2026},
  url          = {https://proceedings.mlsys.org/paper_files/paper/2026/hash/cbc4ab80cd77aa0eb87da062fbcddb46-Abstract-Conference.html}
}

@inproceedings{agrawal2024sarathi,
  author       = {Amey Agrawal and
      Nitin Kedia and
      Ashish Panwar and
      Jayashree Mohan and
      Nipun Kwatra and
      Bhargav Gulavani and
      Alexey Tumanov and
      Ramachandran Ramjee},
  title        = {{Taming Throughput-Latency Tradeoff in LLM Inference with Sarathi-Serve}},
  booktitle    = {USENIX Symposium on Operating Systems Design and Implementation},
  pages        = {117--134},
  year         = {2024},
  url          = {https://www.usenix.org/conference/osdi24/presentation/agrawal}
}

@inproceedings{dai2024deepseekmoe,
  author       = {Damai Dai and
      Chengqi Deng and
      Chenggang Zhao and
      R. X. Xu and
      Huazuo Gao and
      Deli Chen and
      Jiashi Li and
      Wangding Zeng and
      Xingkai Yu and
      Y. Wu and
      Zhenda Xie and
      Y. K. Li and
      Panpan Huang and
      Fuli Luo and
      Chong Ruan and
      Zhifang Sui and
      Wenfeng Liang},
  title        = {{DeepSeekMoE: Towards Ultimate Expert Specialization in Mixture-of-Experts Language Models}},
  booktitle    = {Annual Meeting of the Association for Computational Linguistics},
  pages        = {1280--1297},
  year         = {2024},
  doi          = {10.18653/V1/2024.ACL-LONG.70},
  url          = {https://doi.org/10.18653/v1/2024.acl-long.70}
}

@article{deepseekai2025deepseekv3,
  author       = {{DeepSeek-AI} and others},
  title        = {{DeepSeek-V3 Technical Report}},
  journal      = {arXiv preprint arXiv:2412.19437v2},
  year         = {2024},
  doi          = {10.48550/arXiv.2412.19437},
  url          = {https://arxiv.org/abs/2412.19437v2},
  note         = {Revised February 18, 2025}
}

@inproceedings{du2022glam,
  author       = {Nan Du and
      Yanping Huang and
      Andrew M. Dai and
      Simon Tong and
      Dmitry Lepikhin and
      Yuanzhong Xu and
      Maxim Krikun and
      Yanqi Zhou and
      Adams Wei Yu and
      Orhan Firat and
      Barret Zoph and
      Liam Fedus and
      Maarten P. Bosma and
      Zongwei Zhou and
      Tao Wang and
      Emma Wang and
      Kellie Webster and
      Marie Pellat and
      Kevin Robinson and
      Kathleen Meier-Hellstern and
      Toju Duke and
      Lucas Dixon and
      Kun Zhang and
      Quoc Le and
      Yonghui Wu and
      Zhifeng Chen and
      Claire Cui},
  title        = {{GLaM: Efficient Scaling of Language Models with Mixture-of-Experts}},
  booktitle    = {International Conference on Machine Learning},
  pages        = {5547--5569},
  year         = {2022},
  url          = {https://proceedings.mlr.press/v162/du22c.html}
}

@inproceedings{fang2025klotski,
  author       = {Zhiyuan Fang and
      Yuegui Huang and
      Zicong Hong and
      Yufeng Lyu and
      Wuhui Chen and
      Yue Yu and
      Fan Yu and
      Zibin Zheng},
  title        = {{Klotski: Efficient Mixture-of-Expert Inference via Expert-Aware Multi-Batch Pipeline}},
  booktitle    = {ACM International Conference on Architectural Support for Programming Languages and Operating Systems},
  pages        = {574--588},
  year         = {2025},
  doi          = {10.1145/3676641.3716261},
  url          = {https://doi.org/10.1145/3676641.3716261}
}

@article{fedus2022switch,
  author       = {William Fedus and
      Barret Zoph and
      Noam Shazeer},
  title        = {{Switch Transformers: Scaling to Trillion Parameter Models with Simple and Efficient Sparsity}},
  journal      = {Journal of Machine Learning Research},
  volume       = {23},
  number       = {120},
  pages        = {1--39},
  year         = {2022},
  url          = {https://jmlr.org/papers/v23/21-0998.html}
}

@inproceedings{fu2025areal,
  author       = {Wei Fu and
      Jiaxuan Gao and
      Xujie Shen and
      Chen Zhu and
      Zhiyu Mei and
      Chuyi He and
      Shusheng Xu and
      Guo Wei and
      Jun Mei and
      Jiashu Wang and
      Tongkai Yang and
      Binhang Yuan and
      Yi Wu},
  title        = {{AReaL: A Large-Scale Asynchronous Reinforcement Learning System for Language Reasoning}},
  booktitle    = {Advances in Neural Information Processing Systems},
  volume       = {38},
  pages        = {36256--36282},
  year         = {2025},
  doi          = {10.52202/085713-1218},
  url          = {https://proceedings.neurips.cc/paper_files/paper/2025/hash/33c00862bfa29ac72ecf630a41e19352-Abstract-Conference.html}
}

@inproceedings{gale2023megablocks,
  author       = {Trevor Gale and
      Deepak Narayanan and
      Cliff Young and
      Matei Zaharia},
  title        = {{MegaBlocks: Efficient Sparse Training with Mixture-of-Experts}},
  booktitle    = {Proceedings of Machine Learning and Systems},
  volume       = {5},
  pages        = {288--304},
  year         = {2023},
  url          = {https://proceedings.mlsys.org/paper_files/paper/2023/hash/5a54f79333768effe7e8927bcccffe40-Abstract-mlsys2023.html}
}

@inproceedings{gao2026rollpacker,
  author       = {Wei Gao and
      Yuheng Zhao and
      Dakai An and
      Tianyuan Wu and
      Lunxi Cao and
      Shaopan Xiong and
      Ju Huang and
      Weixun Wang and
      Siran Yang and
      Wenbo Su and
      Jiamang Wang and
      Lin Qu and
      Bo Zheng and
      Wei Wang},
  title        = {{RollPacker: Taming Long-Tail Rollouts for RL Post-Training with Tail Batching}},
  booktitle    = {USENIX Symposium on Networked Systems Design and Implementation},
  pages        = {849--866},
  year         = {2026},
  url          = {https://www.usenix.org/conference/nsdi26/presentation/gao-wei}
}

@article{guo2025deepseekr1,
  author       = {Daya Guo and Dejian Yang and Haowei Zhang and others},
  title        = {{DeepSeek-R1 incentivizes reasoning in LLMs through reinforcement learning}},
  journal      = {Nature},
  volume       = {645},
  number       = {8081},
  pages        = {633--638},
  year         = {2025},
  month        = sep,
  doi          = {10.1038/s41586-025-09422-z},
  url          = {https://doi.org/10.1038/s41586-025-09422-z}
}

@inproceedings{hagberg2008networkx,
  author       = {Aric A. Hagberg and
      Daniel A. Schult and
      Pieter J. Swart},
  title        = {{Exploring Network Structure, Dynamics, and Function using NetworkX}},
  booktitle    = {Proceedings of the 7th Python in Science Conference},
  pages        = {11--15},
  year         = {2008},
  doi          = {10.25080/TCWV9851},
  url          = {https://doi.org/10.25080/TCWV9851}
}

@inproceedings{he2022fastermoe,
  author       = {Jiaao He and
      Jidong Zhai and
      Tiago Antunes and
      Haojie Wang and
      Fuwen Luo and
      Shangfeng Shi and
      Qin Li},
  title        = {{FasterMoE: modeling and optimizing training of large-scale dynamic pre-trained models}},
  booktitle    = {ACM SIGPLAN Symposium on Principles and Practice of Parallel Programming},
  pages        = {120--134},
  year         = {2022},
  doi          = {10.1145/3503221.3508418},
  url          = {https://doi.org/10.1145/3503221.3508418}
}

@inproceedings{he2026rhymerl,
  author       = {Jingkai He and
      Tianjian Li and
      Erhu Feng and
      Dong Du and
      Qian Liu and
      Tao Liu and
      Yubin Xia and
      Haibo Chen},
  title        = {{History Doesn't Repeat Itself but Rollouts Rhyme: Accelerating Reinforcement Learning with RhymeRL}},
  booktitle    = {ACM International Conference on Architectural Support for Programming Languages and Operating Systems},
  pages        = {929--945},
  year         = {2026},
  doi          = {10.1145/3779212.3790172},
  url          = {https://doi.org/10.1145/3779212.3790172}
}

@inproceedings{hu2026tlt,
  author       = {Qinghao Hu and
      Shang Yang and
      Junxian Guo and
      Xiaozhe Yao and
      Yujun Lin and
      Yuxian Gu and
      Han Cai and
      Chuang Gan and
      Ana Klimovic and
      Song Han},
  title        = {{Taming the Long-Tail: Efficient Reasoning RL Training with Adaptive Drafter}},
  booktitle    = {ACM International Conference on Architectural Support for Programming Languages and Operating Systems},
  pages        = {1933--1948},
  year         = {2026},
  doi          = {10.1145/3779212.3790231},
  url          = {https://doi.org/10.1145/3779212.3790231}
}

@article{zeng2025efficientmoe,
  author       = {Yan Zeng and
      Chengchuang Huang and
      Yipeng Mei and
      Lifu Zhang and
      Teng Su and
      Wei Ye and
      Wenqi Shi and
      Shengnan Wang},
  title        = {{EfficientMoE: Optimizing Mixture-of-Experts Model Training With Adaptive Load Balance}},
  journal      = {IEEE Trans. Parallel Distributed Syst.},
  volume       = {36},
  number       = {4},
  pages        = {677--688},
  year         = {2025},
  doi          = {10.1109/TPDS.2025.3539297},
  url          = {https://doi.org/10.1109/TPDS.2025.3539297}
}

@inproceedings{hwang2023tutel,
  author       = {Changho Hwang and
      Wei Cui and
      Yifan Xiong and
      Ziyue Yang and
      Ze Liu and
      Han Hu and
      Zilong Wang and
      Rafael Salas and
      Jithin Jose and
      Prabhat Ram and
      HoYuen Chau and
      Peng Cheng and
      Fan Yang and
      Mao Yang and
      Yongqiang Xiong},
  title        = {{Tutel: Adaptive Mixture-of-Experts at Scale}},
  booktitle    = {Proceedings of Machine Learning and Systems},
  volume       = {5},
  pages        = {269--287},
  year         = {2023},
  url          = {https://proceedings.mlsys.org/paper_files/paper/2023/hash/5616d34cf8ff73942cfd5aa922842556-Abstract-mlsys2023.html}
}

@inproceedings{hwang2024pregated,
  author       = {Ranggi Hwang and
      Jianyu Wei and
      Shijie Cao and
      Changho Hwang and
      Xiaohu Tang and
      Ting Cao and
      Mao Yang},
  title        = {{Pre-gated MoE: An Algorithm-System Co-Design for Fast and Scalable Mixture-of-Expert Inference}},
  booktitle    = {ACM/IEEE International Symposium on Computer Architecture},
  pages        = {1018--1031},
  year         = {2024},
  doi          = {10.1109/ISCA59077.2024.00078},
  url          = {https://doi.org/10.1109/ISCA59077.2024.00078}
}

@article{jiang2024mixtral,
  author       = {Albert Q. Jiang and
      Alexandre Sablayrolles and
      Antoine Roux and
      Arthur Mensch and
      Blanche Savary and
      Chris Bamford and
      Devendra Singh Chaplot and
      Diego de las Casas and
      Emma Bou Hanna and
      Florian Bressand and
      Gianna Lengyel and
      Guillaume Bour and
      Guillaume Lample and
      L{\'e}lio Renard Lavaud and
      Lucile Saulnier and
      Marie-Anne Lachaux and
      Pierre Stock and
      Sandeep Subramanian and
      Sophia Yang and
      Szymon Antoniak and
      Teven Le Scao and
      Th{\'e}ophile Gervet and
      Thibaut Lavril and
      Thomas Wang and
      Timoth{\'e}e Lacroix and
      William El Sayed},
  title        = {{Mixtral of Experts}},
  journal      = {arXiv preprint arXiv:2401.04088},
  year         = {2024},
  doi          = {10.48550/ARXIV.2401.04088},
  url          = {https://doi.org/10.48550/arXiv.2401.04088}
}

@inproceedings{kamahori2025fiddler,
  author       = {Keisuke Kamahori and
      Tian Tang and
      Yile Gu and
      Kan Zhu and
      Baris Kasikci},
  title        = {{Fiddler: CPU-GPU Orchestration for Fast Inference of Mixture-of-Experts Models}},
  booktitle    = {International Conference on Learning Representations},
  year         = {2025},
  url          = {https://proceedings.iclr.cc/paper_files/paper/2025/hash/8cd1ce03ea58b3d7dfd809e4d42f08ea-Abstract-Conference.html}
}

@inproceedings{kwon2023pagedattention,
  author       = {Woosuk Kwon and
      Zhuohan Li and
      Siyuan Zhuang and
      Ying Sheng and
      Lianmin Zheng and
      Cody Hao Yu and
      Joseph Gonzalez and
      Hao Zhang and
      Ion Stoica},
  title        = {{Efficient Memory Management for Large Language Model Serving with PagedAttention}},
  booktitle    = {ACM Symposium on Operating Systems Principles},
  pages        = {611--626},
  year         = {2023},
  doi          = {10.1145/3600006.3613165},
  url          = {https://doi.org/10.1145/3600006.3613165}
}

@inproceedings{lepikhin2021gshard,
  author       = {Dmitry Lepikhin and
      HyoukJoong Lee and
      Yuanzhong Xu and
      Dehao Chen and
      Orhan Firat and
      Yanping Huang and
      Maxim Krikun and
      Noam Shazeer and
      Zhifeng Chen},
  title        = {{GShard: Scaling Giant Models with Conditional Computation and Automatic Sharding}},
  booktitle    = {International Conference on Learning Representations},
  year         = {2021},
  url          = {https://openreview.net/forum?id=qrwe7XHTmYb}
}

@inproceedings{leviathan2023speculative,
  author       = {Yaniv Leviathan and
      Matan Kalman and
      Yossi Matias},
  title        = {{Fast Inference from Transformers via Speculative Decoding}},
  booktitle    = {International Conference on Machine Learning},
  pages        = {19274--19286},
  year         = {2023},
  url          = {https://proceedings.mlr.press/v202/leviathan23a.html}
}

@inproceedings{li2024eagle,
  author       = {Yuhui Li and
      Fangyun Wei and
      Chao Zhang and
      Hongyang Zhang},
  title        = {{EAGLE: Speculative Sampling Requires Rethinking Feature Uncertainty}},
  booktitle    = {International Conference on Machine Learning},
  pages        = {28935--28948},
  year         = {2024},
  url          = {https://proceedings.mlr.press/v235/li24bt.html}
}

@inproceedings{mei2025real,
  author       = {Zhiyu Mei and
      Wei Fu and
      Kaiwei Li and
      Guangju Wang and
      Huanchen Zhang and
      Yi Wu},
  title        = {{ReaL: Efficient RLHF Training of Large Language Models with Parameter Reallocation}},
  booktitle    = {Proceedings of Machine Learning and Systems},
  volume       = {7},
  year         = {2025},
  url          = {https://proceedings.mlsys.org/paper_files/paper/2025/hash/3b3889d313ba9476c12c2d77ea66b24f-Abstract-Conference.html}
}

@inproceedings{miao2024spotserve,
  author       = {Xupeng Miao and
      Chunan Shi and
      Jiangfei Duan and
      Xiaoli Xi and
      Dahua Lin and
      Bin Cui and
      Zhihao Jia},
  title        = {{SpotServe: Serving Generative Large Language Models on Preemptible Instances}},
  booktitle    = {ACM International Conference on Architectural Support for Programming Languages and Operating Systems},
  pages        = {1112--1127},
  year         = {2024},
  doi          = {10.1145/3620665.3640411},
  url          = {https://doi.org/10.1145/3620665.3640411}
}

@article{nie2023flexmoe,
  author       = {Xiaonan Nie and
      Xupeng Miao and
      Zilong Wang and
      Zichao Yang and
      Jilong Xue and
      Lingxiao Ma and
      Gang Cao and
      Bin Cui},
  title        = {{FlexMoE: Scaling Large-scale Sparse Pre-trained Model Training via Dynamic Device Placement}},
  journal      = {Proc. ACM Manag. Data},
  volume       = {1},
  number       = {1},
  pages        = {110:1--110:19},
  year         = {2023},
  doi          = {10.1145/3588964},
  url          = {https://doi.org/10.1145/3588964}
}

@inproceedings{ouyang2022training,
  author       = {Long Ouyang and
      Jeffrey Wu and
      Xu Jiang and
      Diogo Almeida and
      Carroll Wainwright and
      Pamela Mishkin and
      Chong Zhang and
      Sandhini Agarwal and
      Katarina Slama and
      Alex Ray and
      John Schulman and
      Jacob Hilton and
      Fraser Kelton and
      Luke Miller and
      Maddie Simens and
      Amanda Askell and
      Peter Welinder and
      Paul F. Christiano and
      Jan Leike and
      Ryan Lowe},
  title        = {{Training language models to follow instructions with human feedback}},
  booktitle    = {Advances in Neural Information Processing Systems},
  volume       = {35},
  pages        = {27730--27744},
  year         = {2022},
  doi          = {10.52202/068431-2011},
  url          = {https://proceedings.neurips.cc/paper_files/paper/2022/hash/b1efde53be364a73914f58805a001731-Abstract-Conference.html}
}

@inproceedings{prabhu2025vattention,
  author       = {Ramya Prabhu and
      Ajay Nayak and
      Jayashree Mohan and
      Ramachandran Ramjee and
      Ashish Panwar},
  title        = {{vAttention: Dynamic Memory Management for Serving LLMs without PagedAttention}},
  booktitle    = {ACM International Conference on Architectural Support for Programming Languages and Operating Systems},
  pages        = {1133--1150},
  year         = {2025},
  doi          = {10.1145/3669940.3707256},
  url          = {https://doi.org/10.1145/3669940.3707256}
}

@inproceedings{qiao2021pollux,
  author       = {Aurick Qiao and
      Sang Keun Choe and
      Suhas Jayaram Subramanya and
      Willie Neiswanger and
      Qirong Ho and
      Hao Zhang and
      Gregory R. Ganger and
      Eric P. Xing},
  title        = {{Pollux: Co-adaptive Cluster Scheduling for Goodput-Optimized Deep Learning}},
  booktitle    = {USENIX Symposium on Operating Systems Design and Implementation},
  pages        = {1--18},
  year         = {2021},
  url          = {https://www.usenix.org/conference/osdi21/presentation/qiao}
}

@inproceedings{qin2025mooncake,
  author       = {Ruoyu Qin and
      Zheming Li and
      Weiran He and
      Jialei Cui and
      Feng Ren and
      Mingxing Zhang and
      Yongwei Wu and
      Weimin Zheng and
      Xinran Xu},
  title        = {{Mooncake: Trading More Storage for Less Computation - A KVCache-centric Architecture for Serving LLM Chatbot}},
  booktitle    = {USENIX Conference on File and Storage Technologies},
  pages        = {155--170},
  year         = {2025},
  url          = {https://www.usenix.org/conference/fast25/presentation/qin}
}

@inproceedings{rajbhandari2022deepspeedmoe,
  author       = {Samyam Rajbhandari and
      Conglong Li and
      Zhewei Yao and
      Minjia Zhang and
      Reza Yazdani Aminabadi and
      Ammar Ahmad Awan and
      Jeff Rasley and
      Yuxiong He},
  title        = {{DeepSpeed-MoE: Advancing Mixture-of-Experts Inference and Training to Power Next-Generation AI Scale}},
  booktitle    = {International Conference on Machine Learning},
  pages        = {18332--18346},
  year         = {2022},
  url          = {https://proceedings.mlr.press/v162/rajbhandari22a.html}
}

@article{schulman2017ppo,
  author       = {John Schulman and
      Filip Wolski and
      Prafulla Dhariwal and
      Alec Radford and
      Oleg Klimov},
  title        = {{Proximal Policy Optimization Algorithms}},
  journal      = {arXiv preprint arXiv:1707.06347},
  year         = {2017},
  doi          = {10.48550/arXiv.1707.06347},
  url          = {https://arxiv.org/abs/1707.06347}
}

@article{shao2024deepseekmath,
  author       = {Zhihong Shao and
      Peiyi Wang and
      Qihao Zhu and
      Runxin Xu and
      Junxiao Song and
      Xiao Bi and
      Haowei Zhang and
      Mingchuan Zhang and
      Y. K. Li and
      Y. Wu and
      Daya Guo},
  title        = {{DeepSeekMath: Pushing the Limits of Mathematical Reasoning in Open Language Models}},
  journal      = {arXiv preprint arXiv:2402.03300},
  year         = {2024},
  doi          = {10.48550/ARXIV.2402.03300},
  url          = {https://arxiv.org/abs/2402.03300}
}

@inproceedings{shazeer2017sparsely,
  author       = {Noam Shazeer and
      Azalia Mirhoseini and
      Krzysztof Maziarz and
      Andy Davis and
      Quoc Le and
      Geoffrey Hinton and
      Jeff Dean},
  title        = {{Outrageously Large Neural Networks: The Sparsely-Gated Mixture-of-Experts Layer}},
  booktitle    = {International Conference on Learning Representations},
  year         = {2017},
  url          = {https://openreview.net/forum?id=B1ckMDqlg}
}

@inproceedings{sheng2025hybridflow,
  author       = {Guangming Sheng and
      Chi Zhang and
      Zilingfeng Ye and
      Xibin Wu and
      Wang Zhang and
      Ru Zhang and
      Yanghua Peng and
      Haibin Lin and
      Chuan Wu},
  title        = {{HybridFlow: A Flexible and Efficient RLHF Framework}},
  booktitle    = {European Conference on Computer Systems},
  pages        = {1279--1297},
  year         = {2025},
  doi          = {10.1145/3689031.3696075},
  url          = {https://doi.org/10.1145/3689031.3696075}
}

@inproceedings{stiennon2020learning,
  author       = {Nisan Stiennon and
      Long Ouyang and
      Jeffrey Wu and
      Daniel Ziegler and
      Ryan Lowe and
      Chelsea Voss and
      Alec Radford and
      Dario Amodei and
      Paul F. Christiano},
  title        = {{Learning to summarize with human feedback}},
  booktitle    = {Advances in Neural Information Processing Systems},
  volume       = {33},
  pages        = {3008--3021},
  year         = {2020},
  url          = {https://proceedings.neurips.cc/paper/2020/hash/1f89885d556929e98d3ef9b86448f951-Abstract.html}
}

@inproceedings{strati2024orion,
  author       = {Foteini Strati and
      Xianzhe Ma and
      Ana Klimovic},
  title        = {{Orion: Interference-aware, Fine-grained GPU Sharing for ML Applications}},
  booktitle    = {European Conference on Computer Systems},
  pages        = {1075--1092},
  year         = {2024},
  doi          = {10.1145/3627703.3629578},
  url          = {https://doi.org/10.1145/3627703.3629578}
}

@misc{tan2025deepscaler,
  author       = {Michael Luo and
      Sijun Tan and
      Justin Wong and
      Xiaoxiang Shi and
      William Y. Tang and
      Manan Roongta and
      Colin Cai and
      Jeffrey Luo and
      Li Erran Li and
      Raluca Ada Popa and
      Ion Stoica},
  title        = {{DeepScaleR: Surpassing O1-Preview with a 1.5B Model by Scaling RL}},
  year         = {2025},
  month        = feb,
  howpublished = {Agentica project report},
  url          = {https://pretty-radio-b75.notion.site/DeepScaleR-Surpassing-O1-Preview-with-a-1-5B-Model-by-Scaling-RL-19681902c1468005bed8ca303013a4e2}
}

@article{tang2025pangu,
  author       = {Yehui Tang and
      Xiaosong Li and
      Fangcheng Liu and
      Wei Guo and
      Hang Zhou and
      Yaoyuan Wang and
      Kai Han and
      Xianzhi Yu and
      Jinpeng Li and
      Hui Zang and
      Fei Mi and
      Xiaojun Meng and
      Zhicheng Liu and
      Hanting Chen and
      Binfan Zheng and
      Can Chen and
      Youliang Yan and
      Ruiming Tang and
      Peifeng Qin and
      Xinghao Chen and
      Dacheng Tao and
      Yunhe Wang and {Other Contributors}},
  title        = {{Pangu Pro MoE: Mixture of Grouped Experts for Efficient Sparsity}},
  journal      = {arXiv preprint arXiv:2505.21411},
  year         = {2025},
  doi          = {10.48550/ARXIV.2505.21411},
  url          = {https://doi.org/10.48550/arXiv.2505.21411}
}

@article{virtanen2020scipy,
  author       = {Pauli Virtanen and
      Ralf Gommers and
      Travis E. Oliphant and
      Matt Haberland and
      Tyler Reddy and
      David Cournapeau and
      Evgeni Burovski and
      Pearu Peterson and
      Warren Weckesser and
      Jonathan Bright and
      St{\'e}fan J. van der Walt and
      Matthew Brett and
      Joshua Wilson and
      K. Jarrod Millman and
      Nikolay Mayorov and
      Andrew R. J. Nelson and
      Eric Jones and
      Robert Kern and
      Eric Larson and
      C. J. Carey and
      {\.{I}}lhan Polat and
      Yu Feng and
      Eric W. Moore and
      Jake VanderPlas and
      Denis Laxalde and
      Josef Perktold and
      Robert Cimrman and
      Ian Henriksen and
      E. A. Quintero and
      Charles R. Harris and
      Anne M. Archibald and
      Ant{\^o}nio H. Ribeiro and
      Fabian Pedregosa and
      Paul van Mulbregt and {{SciPy 1.0 Contributors}}},
  title        = {{SciPy 1.0: Fundamental Algorithms for Scientific Computing in Python}},
  journal      = {Nature Methods},
  volume       = {17},
  number       = {3},
  pages        = {261--272},
  year         = {2020},
  doi          = {10.1038/s41592-019-0686-2},
  url          = {https://doi.org/10.1038/s41592-019-0686-2}
}

@inproceedings{wang2025sirius,
  author       = {Jiali Wang and
      Yankui Wang and
      Mingcong Han and
      Rong Chen},
  title        = {{Colocating ML Inference and Training with Fast GPU Memory Handover}},
  booktitle    = {USENIX Annual Technical Conference},
  pages        = {1657--1675},
  year         = {2025},
  url          = {https://www.usenix.org/conference/atc25/presentation/wang-jiali}
}

@inproceedings{yu2025dapo,
  author       = {Qiying Yu and
      Zheng Zhang and
      Ruofei Zhu and
      Yufeng Yuan and
      Xiaochen Zuo and
      Yu Yue and
      Weinan Dai and
      Tiantian Fan and
      Gaohong Liu and
      Juncai Liu and
      Lingjun Liu and
      Xin Liu and
      Haibin Lin and
      Zhiqi Lin and
      Bole Ma and
      Guangming Sheng and
      Yuxuan Tong and
      Chi Zhang and
      Mofan Zhang and
      Ru Zhang and
      Wang Zhang and
      Hang Zhu and
      Jinhua Zhu and
      Jiaze Chen and
      Jiangjie Chen and
      Chengyi Wang and
      Hongli Yu and
      Yuxuan Song and
      Xiangpeng Wei and
      Hao Zhou and
      Jingjing Liu and
      Wei-Ying Ma and
      Ya-Qin Zhang and
      Lin Yan and
      Yonghui Wu and
      Mingxuan Wang},
  title        = {{DAPO: An Open-Source LLM Reinforcement Learning System at Scale}},
  booktitle    = {Advances in Neural Information Processing Systems},
  volume       = {38},
  pages        = {113222--113244},
  year         = {2025},
  doi          = {10.52202/085713-3775},
  url          = {https://proceedings.neurips.cc/paper_files/paper/2025/hash/a4277440d50f1f15d2cb4c14f7e0c0d2-Abstract-Conference.html}
}

@inproceedings{zhai2023smartmoe,
  author       = {Mingshu Zhai and
      Jiaao He and
      Zixuan Ma and
      Zan Zong and
      Runqing Zhang and
      Jidong Zhai},
  title        = {{SmartMoE: Efficiently Training Sparsely-Activated Models through Combining Offline and Online Parallelization}},
  booktitle    = {USENIX Annual Technical Conference},
  pages        = {961--975},
  year         = {2023},
  url          = {https://www.usenix.org/conference/atc23/presentation/zhai}
}

@inproceedings{zheng2024sglang,
  author       = {Lianmin Zheng and
      Liangsheng Yin and
      Zhiqiang Xie and
      Chuyue Sun and
      Jeff Huang and
      Cody Hao Yu and
      Shiyi Cao and
      Christos Kozyrakis and
      Ion Stoica and
      Joseph E. Gonzalez and
      Clark Barrett and
      Ying Sheng},
  title        = {{SGLang: Efficient Execution of Structured Language Model Programs}},
  booktitle    = {Advances in Neural Information Processing Systems},
  volume       = {37},
  pages        = {62557--62583},
  year         = {2024},
  doi          = {10.52202/079017-2000},
  url          = {https://proceedings.neurips.cc/paper_files/paper/2024/hash/724be4472168f31ba1c9ac630f15dec8-Abstract-Conference.html}
}

@inproceedings{zhong2024distserve,
  author       = {Yinmin Zhong and
      Shengyu Liu and
      Junda Chen and
      Jianbo Hu and
      Yibo Zhu and
      Xuanzhe Liu and
      Xin Jin and
      Hao Zhang},
  title        = {{DistServe: Disaggregating Prefill and Decoding for Goodput-optimized Large Language Model Serving}},
  booktitle    = {USENIX Symposium on Operating Systems Design and Implementation},
  pages        = {193--210},
  year         = {2024},
  url          = {https://www.usenix.org/conference/osdi24/presentation/zhong-yinmin}
}

@inproceedings{zhong2025rlhfuse,
  author       = {Yinmin Zhong and
      Zili Zhang and
      Bingyang Wu and
      Shengyu Liu and
      Yukun Chen and
      Changyi Wan and
      Hanpeng Hu and
      Lei Xia and
      Ranchen Ming and
      Yibo Zhu and
      Xin Jin},
  title        = {{Optimizing RLHF Training for Large Language Models with Stage Fusion}},
  booktitle    = {USENIX Symposium on Networked Systems Design and Implementation},
  pages        = {489--503},
  year         = {2025},
  url          = {https://www.usenix.org/conference/nsdi25/presentation/zhong}
}

@article{ziegler2019fine,
  author       = {Daniel M. Ziegler and
      Nisan Stiennon and
      Jeffrey Wu and
      Tom B. Brown and
      Alec Radford and
      Dario Amodei and
      Paul Christiano and
      Geoffrey Irving},
  title        = {{Fine-Tuning Language Models from Human Preferences}},
  journal      = {arXiv preprint arXiv:1909.08593},
  year         = {2019},
  doi          = {10.48550/arXiv.1909.08593},
  url          = {https://arxiv.org/abs/1909.08593}
}

@article{cobbe2021gsm8k,
  author       = {Karl Cobbe and Vineet Kosaraju and Mohammad Bavarian and
                  Mark Chen and Heewoo Jun and Lukasz Kaiser and
                  Matthias Plappert and Jerry Tworek and Jacob Hilton and
                  Reiichiro Nakano and Christopher Hesse and John Schulman},
  title        = {{Training Verifiers to Solve Math Word Problems}},
  journal      = {arXiv preprint arXiv:2110.14168v2},
  year         = {2021},
  doi          = {10.48550/arXiv.2110.14168},
  url          = {https://arxiv.org/abs/2110.14168v2}
}

@article{cui2025prime,
  author       = {Ganqu Cui and
                  Lifan Yuan and
                  Zefan Wang and
                  Hanbin Wang and
                  Yuchen Zhang and
                  Jiacheng Chen and
                  Wendi Li and
                  Bingxiang He and
                  Yuchen Fan and
                  Tianyu Yu and
                  Qixin Xu and
                  Weize Chen and
                  Jiarui Yuan and
                  Huayu Chen and
                  Kaiyan Zhang and
                  Xingtai Lv and
                  Shuo Wang and
                  Yuan Yao and
                  Xu Han and
                  Hao Peng and
                  Yu Cheng and
                  Zhiyuan Liu and
                  Maosong Sun and
                  Bowen Zhou and
                  Ning Ding},
  title        = {{Process Reinforcement through Implicit Rewards}},
  journal      = {Transactions on Machine Learning Research},
  year         = {2026},
  month        = may,
  url          = {https://openreview.net/forum?id=9SkkifLopZ}
}

@inproceedings{zhu2025megascaleinfer,
  author       = {Ruidong Zhu and Ziheng Jiang and Chao Jin and Peng Wu and
      Cesar A. Stuardo and Dongyang Wang and Xinlei Zhang and Huaping Zhou and
      Haoran Wei and Yang Cheng and Jianzhe Xiao and Xinyi Zhang and
      Lingjun Liu and Haibin Lin and Li-Wen Chang and Jianxi Ye and Xiao Yu and
      Xuanzhe Liu and Xin Jin and Xin Liu},
  title        = {{MegaScale-Infer: Efficient Mixture-of-Experts Model Serving with Disaggregated Expert Parallelism}},
  booktitle    = {Proceedings of the ACM SIGCOMM 2025 Conference},
  pages        = {592--608},
  year         = {2025},
  doi          = {10.1145/3718958.3750506},
  url          = {https://doi.org/10.1145/3718958.3750506}
}

@article{wang2026ainferpd,
  author       = {Guowei Wang and Chaokun Yang and Zhenxuan Pan and Yipeng Wei and
      Yuhong Guo and Minghua Zhu and Zechuan Zhang and Shuo Wan and Xiaowei Zhu},
  title        = {{AInfer-PD: Communication-Safe In-Place Prefill-Decode Multiplexing for Distributed MoE Rollouts}},
  journal      = {arXiv preprint arXiv:2609.00993v3},
  year         = {2026},
  doi          = {10.48550/arXiv.2609.00993},
  url          = {https://arxiv.org/abs/2609.00993v3}
}

@article{roberts1959geometric,
  author       = {S. W. Roberts},
  title        = {Control Chart Tests Based on Geometric Moving Averages},
  journal      = {Technometrics},
  volume       = {1},
  number       = {3},
  pages        = {239--250},
  year         = {1959},
  doi          = {10.1080/00401706.1959.10489860},
  url          = {https://doi.org/10.1080/00401706.1959.10489860}
}

\clearpage

\appendix
\section{Device Memory Across Policy Updates}

\begin{figure}[!ht]
  \centering
  \includegraphics[width=0.92\linewidth]{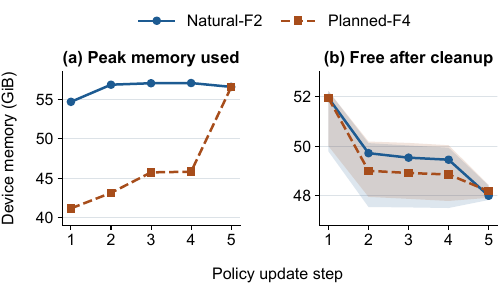}
  \caption{Device memory across the first five policy updates. The left panel
  shows the maximum memory used across sampled phases and ranks. The right
  panel shows median free memory after rollout release and cache cleanup,
  with bands spanning all 16 ranks.}
  \label{fig:supplement-memory}
\end{figure}

Figure~\ref{fig:supplement-memory} gives the step-by-step memory measurements behind the
``Memory feasibility and reclamation'' paragraph in the paper.
Natural-F2 and Planned-F4 both completed five consecutive policy updates
without running out of memory, aborting responses, or triggering KV
preemption. The median increase in free device memory from before local
rollout release to after cache cleanup was 42.98~GiB for Natural-F2 and
32.09~GiB for Planned-F4.

\section{Quota-Aware Prompt Pairing}

Table~\ref{tab:supplement-quota} gives the measurements behind the controlled
comparison in the paper. Both methods generate the same number of response tokens.

\begin{table}[!ht]
  \centering
  \caption{Controlled comparison of MinSkew and quota-aware pairing. Waiting
  runs from local completion to the start of rank exit, or to generation end
  for ranks that remain active.}
  \label{tab:supplement-quota}
  \small
    \setlength{\tabcolsep}{4pt}
  \begin{tabular}{lrr}
    \toprule
    Metric & MinSkew & Quota-aware \\
    \midrule
    Window time (s) &392.41 &385.82 \\
    Throughput (tokens/s) &4358.73 &4433.20 \\
    Occupied rank-time (rank-s) &4791.98 &4458.97 \\
    Resource efficiency (tokens/rank-s) &356.93 &383.59 \\
    Waiting (rank-s) &750.26 &290.65 \\
    Enough ranks finish to reach EP8 (s) &229.56 &194.29 \\
    EP8 transition call ends (s) &245.68 &210.41 \\
    Enough ranks finish to reach EP4 (s) &293.41 &287.43 \\
    EP4 transition call ends (s) &308.59 &303.11 \\
    \bottomrule
  \end{tabular}
\end{table}
\end{document}